\documentclass{article}

\usepackage{PRIMEarxiv}

\usepackage[utf8]{inputenc} % allow utf-8 input
\usepackage[T1]{fontenc}    % use 8-bit T1 fonts
\usepackage{hyperref}       % hyperlinks
\usepackage{url}            % simple URL typesetting
\usepackage{booktabs}       % professional-quality tables
\usepackage{amsfonts}       % blackboard math symbols
\usepackage{nicefrac}       % compact symbols for 1/2, etc.
\usepackage{microtype}      % microtypography
\usepackage{lipsum}
\usepackage{fancyhdr}       % header
\usepackage{graphicx}       % graphics
\graphicspath{{media/}}     % organize your images and other figures under media/ folder

\usepackage{adjustbox}
\usepackage{bbm}
\usepackage{bm}
\usepackage{amsmath}
\usepackage{booktabs}

\usepackage{makecell}
\usepackage{caption}
\usepackage{subcaption}
\usepackage{multirow}
\usepackage{xurl}
\usepackage[dvipsnames]{xcolor}

\usepackage{microtype}

\usepackage{array}
\usepackage{enumitem}
\usepackage{bm}
\usepackage{amsmath}
\usepackage{booktabs}
\usepackage{algorithm}
\usepackage{algpseudocode}
\usepackage{amssymb}
\usepackage{pifont}
\usepackage{tabularx}
\newcolumntype{Y}{>{\centering\arraybackslash}X}

\usepackage{cleveref}
\newcommand{\xmark}{\ding{55}} % Define ✗
\newcommand{\cmark}{\ding{51}} % Define ✓
\usepackage{multirow}

\title{\Large\bfseries Descrip3D: Enhancing Large Language Model-based 3D Scene Understanding with Object-Level Text Descriptions}

\usepackage{authblk}
\author[1]{Jintang Xue}
\author[1]{Ganning Zhao}
\author[1]{Jie-En Yao}
\author[1]{Hong-En Chen}
\author[1]{Yue Hu}
\author[1]{Meida Chen}
\author[2]{Suya You}
\author[1]{C.-C.~Jay~Kuo}
\affil[1]{University of Southern California, Los Angeles, California, USA}
\affil[2]{DEVCOM Army Research Laboratory, Los Angeles, California, USA}

\begin{document}
\maketitle

\begin{abstract}

Understanding 3D scenes goes beyond simply recognizing objects; it requires reasoning about the spatial and semantic relationships between them. Current 3D scene-language models often struggle with this relational understanding, particularly when visual embeddings alone do not adequately convey the roles and interactions of objects. In this paper, we introduce Descrip3D, a novel and powerful framework that explicitly encodes the relationships between objects using natural language. Unlike previous methods that rely only on 2D and 3D embeddings, Descrip3D enhances each object with a textual description that captures both its intrinsic attributes and contextual relationships. These relational cues are incorporated into the model through a dual-level integration: embedding fusion and prompt-level injection. This allows for unified reasoning across various tasks such as grounding, captioning, and question answering, all without the need for task-specific heads or additional supervision. When evaluated on five benchmark datasets, including ScanRefer, Multi3DRefer, ScanQA, SQA3D, and Scan2Cap, Descrip3D consistently outperforms strong baseline models, demonstrating the effectiveness of language-guided relational representation for understanding complex indoor scenes. Our code and data are publicly available at https://github.com/jintangxue/Descrip3D.

\end{abstract}

\section{Introduction}
\label{sec:intro}

\begin{figure}[ht]
  \centering
  \includegraphics[width=0.95\linewidth]{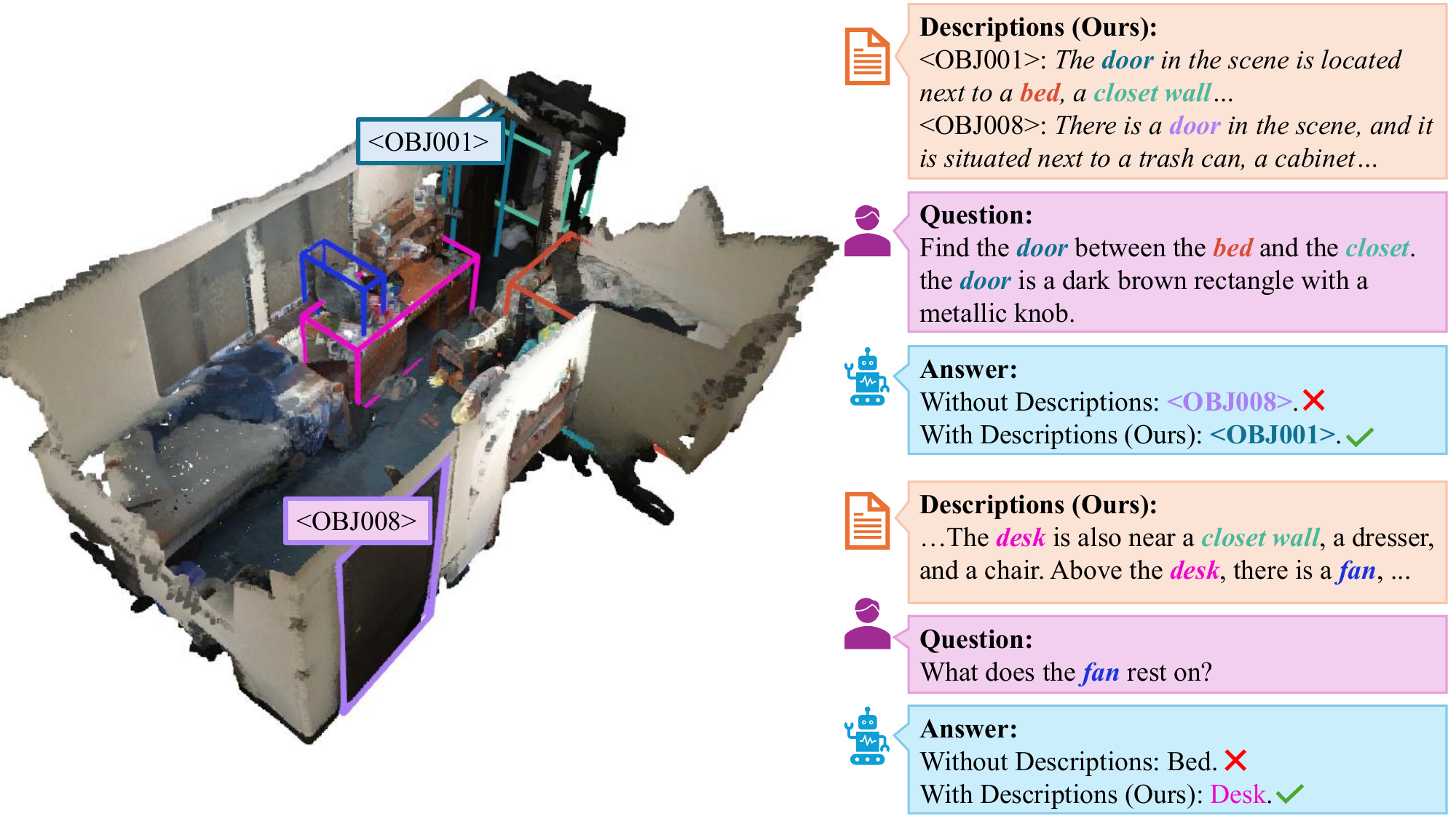}
  \caption{An example of injecting object-level text descriptions during conversation. Providing these descriptions significantly improves the model’s accuracy and reasoning performance.}
  \label{fig:intro}
\end{figure}

Recent advances in large language models (LLMs) have significantly transformed human-computer interaction by equipping machines with unprecedented capabilities in language understanding, reasoning, and open-ended dialogue. LLMs have achieved strong results in domains such as retrieval, code, and multi-modal learning~\cite{hadi2023survey, minaee2024large, zhao2023survey, li2024survey, xue2024bias, wang2023overview}. Building on this progress, researchers have begun to extend LLMs to 3D scene understanding~\cite{ma2024llms, zha2025enable, qi2025gpt4scene}, aiming to empower models with the ability to interpret and reason about complex visual and spatial contexts within real-world environments. In particular, indoor scenes are dense and ambiguous. While early models~\cite{chen2020scanrefer, chen2021scan2cap} targeted single tasks and recent LLMs~\cite{hong20233d, huang2024chat} integrate 2D/3D features, none explicitly model inter-object relations.

As illustrated in Figure~\ref{fig:intro}, we address this limitation by introducing a novel text-based relational modality into the 3D scene language pipeline. Specifically, we augment each object with a natural language description that captures both its intrinsic attributes (e.g., color, material) and its contextual relationships to nearby objects (e.g., 'next to the table', 'under the chair'), generated by prompting a vision-language model with multi-view images. We use a dual-level integration strategy to incorporate the relational descriptions into our model. These descriptions are embedded using a lightweight text encoder and fused with the object's visual embeddings to create a unified multi-modal representation. In addition to embedding-level fusion, the descriptions are also injected into the prompt to support relational understanding in downstream tasks.

This dual-level integration, at both the representation and the language interface levels, allows the model to leverage the prior knowledge of the LLM about the spatial and functional relations learned from the language. Our approach requires no architectural changes, task-specific heads, or extra annotations, making it modular and easy to integrate.

Empirically, our method yields consistent gains across five standard 3D scene language benchmarks. ScanRefer~\cite{chen2020scanrefer}, Multi3DRefer~\cite{zhang2023multi3drefer}, Scan2Cap~\cite{chen2021scan2cap}, ScanQA~\cite{azuma2022scanqa}, and SQA3D~\cite{ma2022sqa3d}. Improvements are especially pronounced in tasks involving complex grounding or multi-object reasoning, highlighting the benefit of explicit text-based relational cues in enhancing 3D scene understanding. Here, multi-object reasoning does not refer only to Multi3DRefer’s multi-target grounding protocol. We use the term more generally to mean queries whose solution depends on object–object relations (e.g., left of, behind, next to, supporting, inside), regardless of whether the benchmark requires multiple boxes.

Our contributions are as follows.
\begin{itemize}
    \item We identify and address a fundamental limitation in existing 3D scene-language models: the absence of explicit modeling of inter-object relationships, which hinders multi-object spatial and semantic reasoning.
    \item We introduce a novel object-centric textual modality that encodes both intrinsic attributes and contextual relationships of objects through automatically generated natural language descriptions.
    \item We propose a dual-level integration framework that incorporates these descriptions both at the embedding fusion level and within language prompts, enabling seamless and architecture-agnostic enhancement of existing LLM-based pipelines.
    \item We validate our approach across five standard 3D scene-language benchmarks, consistently surpassing both expert-designed and LLM-based baselines, especially in tasks demanding relational understanding and compositional reasoning.
\end{itemize}

The remainder of this paper is organized as follows. Section~\ref{sec:related} reviews related work. Section~\ref{sec:method} details our proposed approach. Section~\ref{sec:experiments} presents the experimental results and analysis. Finally, Section~\ref{sec:conclusion} concludes the paper.

\section{Related Work}
\label{sec:related}

\subsection{LLMs for 3D Scene Understanding} 
Research in 3D vision-language has progressed from task-specific solutions to unified and LLM-driven frameworks. Earlier work focused on single-task models for visual grounding~\cite{liu2024survey}, dense  captioning~\cite{yu2023comprehensive}, and question answering~\cite{li2025embodied}. For example, ScanRefer~\cite{chen2020scanrefer}, ReferIt3D~\cite{achlioptas2020referit3d}, and Multi3DRefer~\cite{zhang2023multi3drefer} tackled visual grounding by linking natural language to 3D objects. For dense captioning, 3DVG-Transformer~\cite{zhao20213dvg} and Vote2Cap-DETR++~\cite{chen2024vote2cap} localized and described objects, while ScanQA~\cite{azuma2022scanqa} and SQA3D~\cite{ma2022sqa3d} broadened the scope to open-ended scene-level QA. Building on these, multi-task frameworks such as 3DJCG~\cite{cai20223djcg} and D3Net~\cite{chen2022d3net} unified grounding and captioning with shared encoders, and large-scale pretraining approaches like 3D-VisTA~\cite{zhu20233d} and 3D-VLP~\cite{jin2023context} established transferable 3D vision-language representations. Yet most treat language as auxiliary, limiting explicit relational modeling.

The emergence of large language models (LLMs) has introduced stronger reasoning and conversational capabilities, motivating their integration with 3D data. Approaches such as PointLLM~\cite{xu2024pointllm}, ImageBind-LLM~\cite{han2023imagebind}, Point-Bind~\cite{guo2023point}, Ulip~\cite{xue2023ulip}, and CG3D~\cite{hegde2023clip} align point cloud embeddings with text, enabling object-level captioning and QA. However, these mainly operate on isolated objects, lacking mechanisms for modeling interactions within complex scenes. To overcome this, multimodal LLMs (MLLMs) have been extended to scene-level reasoning. 3D-LLM~\cite{hong20233d} projects 3D embeddings into language space for basic multimodal reasoning, while Chat-3D~\cite{wang2023chat} introduces object-centric prompts and staged training to align 3D objects with dialogue. More recently, Chat-Scene~\cite{huang2024chat} incorporates object identifiers and fuses them with 2D and 3D embeddings, achieving strong results in QA and grounding benchmarks. Yet, even these advances fall short of explicitly encoding inter-object relationships, leaving relational and spatial reasoning an open challenge for LLM-based 3D scene understanding.

\subsection{Explicit Object Relationship Representation}
Although prior models lack explicit mechanisms for encoding inter-object relationships, recent efforts have begun to address this limitation through graph-based or language-based modeling. ConceptGraph~\cite{gu2024conceptgraphs} constructs scene-level concept graphs by aligning 3D object embeddings with textual descriptions using vision-language models, enabling semantic relationship modeling and open-vocabulary generalization. HOV-SG~\cite{werby2024hierarchical} advances this direction by generating holistic object-scene graphs from 3D scans, capturing spatial and semantic relations across the scene. However, graph-based representations require long sequences and incur high inference costs. To address this, 3DGraphLLM~\cite{zemskova20243dgraphllm} incorporates knowledge graph embeddings into object representations, encoding pairwise relationships to support relational reasoning. Yet, this approach introduces significant combinatorial overhead by enumerating object-object triplets and suffers from error propagation in graph construction. Notably, the addition of explicit graph structures in 3DGraphLLM leads to a performance drop on downstream question answering tasks, suggesting that current graph-based formulations may hinder rather than help relational reasoning in practice. Scene-LLM~\cite{fu2024scene} instead leverages LLMs with scene-level prompts to generate global summaries and captions, but lacks fine-grained object-level relational modeling. Thus, explicitly and efficiently representing object relationships in 3D scenes remains an open and challenging problem.

\subsection{3D Dataset Construction via 2D VLMs}
Recent works leverage 2D vision-language models to expand 3D–text resources. Cap3D~\cite{luo2023scalable} generates captions for millions of Objaverse objects by aggregating multi-view 2D renderings with pretrained captioning and alignment models. LiDAR-LLM~\cite{yang2025lidar} reformulates outdoor LiDAR understanding as a language modeling task, producing hundreds of thousands of LiDAR–text pairs through a staged pipeline. These efforts primarily focus on large-scale data creation to support 3D–language research. In contrast, Descrip3D does not rely on new annotations; instead, it injects object-level descriptions directly into the model pipeline to strengthen relational reasoning across existing benchmarks.

\section{Method}
\label{sec:method}

%-------------------------------------------------------------------------
\subsection{Overview}
\begin{figure*}[ht]
  \centering
  \includegraphics[width=0.9\linewidth]{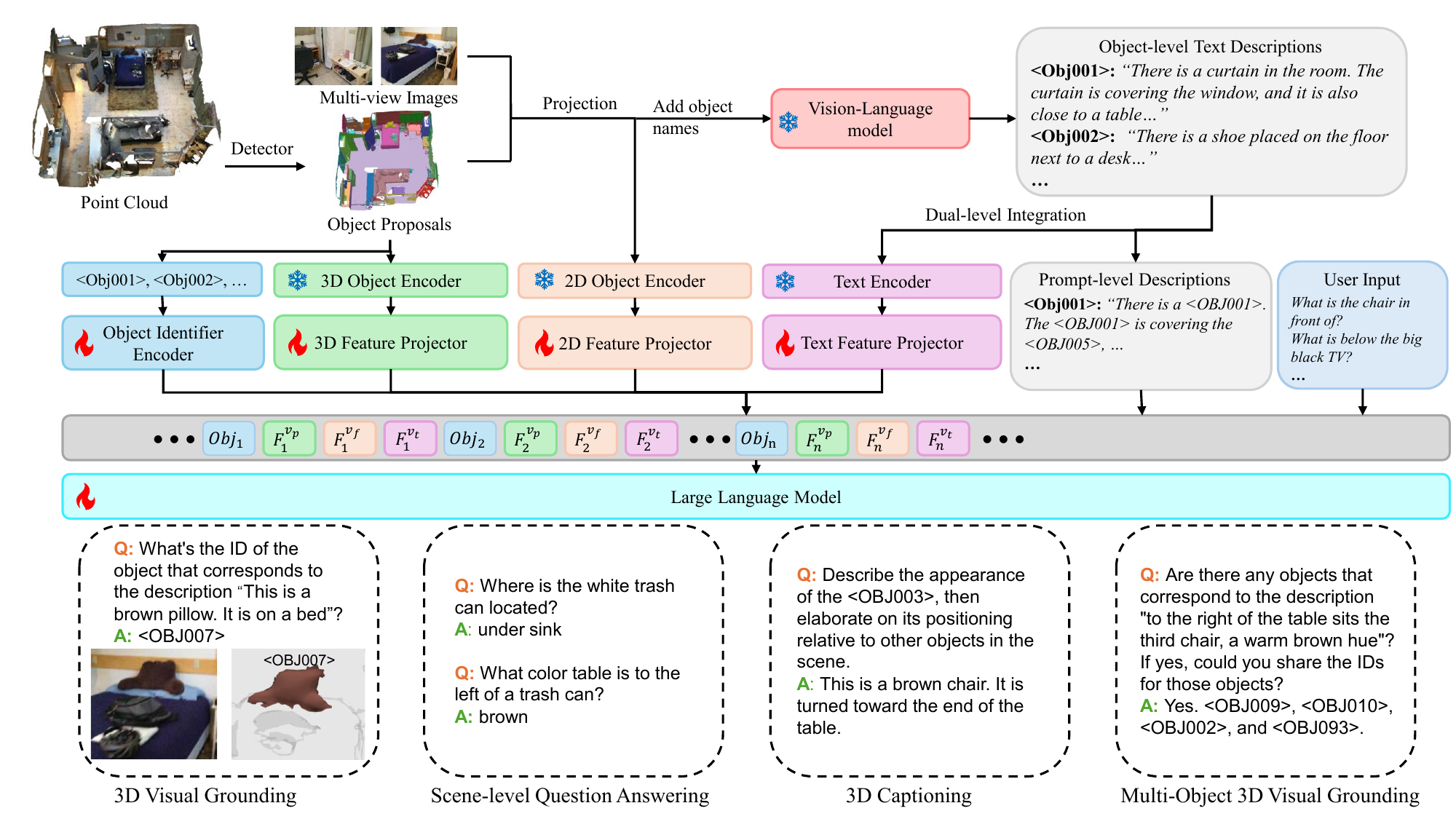}
  \caption{\textbf{Overall model architecture.} We propose a novel and powerful method that explicitly models inter-object relationships by integrating relational text descriptions into object-centric scene representations via a dual-level strategy. From a 3D scan, we extract object proposals and encode their geometry and appearance using pretrained 2D and 3D encoders. Each object is enriched with a natural language description capturing both intrinsic attributes and spatial relations to nearby objects. These descriptions guide scene understanding through: (1) embedding-level fusion with visual features to enhance object representations, and (2) prompt-level injection of queried object descriptions to enhance object-specific relational reasoning. The resulting multimodal tokens enable high-level reasoning for 3D grounding, dense captioning, and question answering. Our design equips the model with both localized and contextual spatial semantics, significantly improving relational reasoning.}
  \label{fig:pipeline}
\end{figure*}

Our method empowers LLMs to perform precise and context-aware reasoning over complex 3D scenes by augmenting object representations with rich, relational textual descriptions. We introduce a powerful text-based relational modality that explicitly encodes both spatial and semantic relationships between objects. Our framework integrates pre-trained 3D and 2D visual encoders, a text encoder, and an LLM, and represents each object as a fused token embedding that combines geometric structure, visual appearance, and relational context. These relational cues are injected into both the object-level embeddings and directly into the LLM prompt, enabling fine-grained multimodal understanding and interaction. This dual-level integration significantly improves the ability of LLM to reason about object relationships, supporting downstream tasks such as grounding, captioning, and question answering with high accuracy and interpretability. An overview of our pipeline is shown in \cref{fig:pipeline}.

%-------------------------------------------------------------------------
\subsection{Object-Centric Scene Representation}
We decompose each scene into discrete object proposals using Mask3D~\cite{schult2022mask3d} first, producing a set of segmented object point clouds $\{\mathbf{P}_1, ..., \mathbf{P}_n\}$. Each object $\mathbf{P}_i \in \mathbb{R}^{m_i \times 6}$ includes XYZ coordinates and RGB values. Here, \(m_i\) refers to the number of points in the proposal of the \(i\)th object. Each resulting object is treated as a fundamental unit for understanding the scene. To support unambiguous reference and interaction, we adopt the design introduced in Chat-Scene~\cite{huang2024chat} to associate each object with a unique learnable identifier token $<\!OBJ_i\!>$. Concretely, each identifier token is inserted into the tokenizer vocabulary as a learnable lexical token and optimized end-to-end.

%-------------------------------------------------------------------------
\subsection{Visual Embedding Extraction} 
For each detected object, we extract geometric and visual embeddings from 3D point clouds and multi-view images. For 3D geometry, we use the pre-trained Uni3D~\cite{zhou2023uni3d} encoder, which produces an embedding $\mathbf{Z}^{v_p}_i \in \mathbb{R}^{1 \times d}$ from each segmented point cloud $\mathbf{P}_i$, and is known for capturing fine-grained structure across diverse scenes. For 2D appearance, we adopt the pre-trained DINOv2~\cite{oquab2023dinov2} vision transformer to extract embeddings from multi-view images. We project each object’s 3D mask onto multi-view images and average the cropped DINOv2 embeddings per view. The features from different views are then aggregated using a size-weighted average into a 2D embedding $\mathbf{Z}^{v_f}_i \in \mathbb{R}^{1 \times d}$. DINOv2 is chosen for its ability to extract semantically rich embeddings from high-resolution input.
%-------------------------------------------------------------------------
\subsection{Text-Based Relational Modality}
While the embeddings extracted from both 2D multi-view images and 3D point clouds are powerful, they are inherently localized to individual objects. 3D embeddings capture geometric and spatial attributes in isolation, while 2D embeddings, though derived from full images, are aggregated per object and provide only limited relational context, making their relational awareness implicit and limited.

As a result, the visual representation lacks explicit modeling of inter-object interactions, limiting its capacity for comprehensive scene understanding and high-level reasoning. To overcome this limitation, we introduce a novel textual modality that encodes relational information in natural language. These descriptions capture how each object is situated within the scene and how it interacts with others, injecting contextual knowledge often inaccessible to vision-based embeddings alone.

Our approach integrates this modality through a dual-level integration: (1) by encoding and fusing text descriptions into object representations and (2) by injecting them into the prompt space of the language model for enhanced reasoning. This design allows our model to exploit both low-level visual embeddings and high-level relational cues, ultimately improving performance on tasks requiring contextual understanding.

\paragraph{Relational Description Generation.} We generate natural language descriptions for each object by utilizing a vision-language model and apply it to multi-view RGB images. For each image, we project 3D object masks onto the 2D plane. This allows us to identify key objects located in the central area of the view, as well as all visible objects in each image. To ensure that the model correctly grounds each object, we overlay the object names (e.g., "couch", "shelf") on the image at their projected centers. We then ask the vision-language model to describe the relationship between each key object and all other visible objects in the image. This results in descriptions at the object level such as: \textit{ ``There is a curtain in the room. The curtain is covering the window, and it is also close to a table.''}

We generate one description per object identifier, skipping any object that has already been described in a previous view to enhance efficiency. In this way, each object receives a single textual description when it appears at the center of at least one multi-view image, resulting in a complete set of object-level relational descriptions for the scene.

\paragraph{Relational Descriptions Encoding.} To encode the generated relational descriptions into the model, we use a Sentence-Transformer model~\cite{reimers-2019-sentence-bert} to convert each object's description into a fixed-dimensional embedding vector $\mathbf{Z}^{v_t}_i$. This representation is subsequently fused into the object representation, complementing the 2D visual embeddings $\mathbf{Z}^{v_f}_i$ and the 3D geometric embeddings $\mathbf{Z}^{v_p}_i$.

\paragraph{Relational Augmented Prompting} In addition to token-level fusion, we leverage a prompt-level strategy that delivers relational priors directly to the language model. When a query mentions an object by name and contains a relational cue (e.g., “next to”), we append a brief description of the relevant object to the end of the input prompt. To avoid semantic ambiguity and ensure consistent grounding, we replace all object names within the descriptions with their corresponding object identifiers. These object identifiers are learned alongside the language model during training, allowing the LLM to associate them with specific objects in the scene. This strategy eliminates confusion caused by duplicate object names (for example, multiple 'chairs') and ensures that relational signals remain precisely anchored to the intended objects. The prompt-level text injection allows the LLM to access relevant contextual cues before reading the user's question, improving reasoning performance.

%-------------------------------------------------------------------------
\subsection{Token Construction and Training Strategy}

To enable unified reasoning across geometry, appearance, and language, all three modalities are projected into a shared token space compatible with the language model. Specifically, we employ separate linear projection heads $f_p$, $f_v$, and $f_t$ to map the original embeddings $\mathbf{Z}^{v_p}_i$, $\mathbf{Z}^{v_f}_i$, and $\mathbf{Z}^{v_t}_i$ into token embeddings $\mathbf{F}^{v_p}_i$, $\mathbf{F}^{v_f}_i$, and $\mathbf{F}^{v_t}_i$:
\begin{align}
\mathbf{F}^{v_p}_i &= f_p(\mathbf{Z}^{v_p}_i), &
\mathbf{F}^{v_f}_i &= f_v(\mathbf{Z}^{v_f}_i), &
\mathbf{F}^{v_t}_i &= f_t(\mathbf{Z}^{v_t}_i).
\end{align}
Each object is represented by a fused token embedding constructed by concatenating its learned identifier token, its 3D geometry embeddings, 2D appearance embeddings, and textual description embeddings as
\begin{equation}
\mathbf{F}_i = \texttt{Concat}(\mathbf{OBJ_i}, \mathbf{F}^{v_p}_i, \mathbf{F}^{v_f}_i, \mathbf{F}^{v_t}_i).
\end{equation}
These object tokens are serialized into the language model input in the format:
[\textless OBJ001\textgreater $\mathbf{F}_1$, \textless OBJ002\textgreater $\mathbf{F}_2$, ..., \textless OBJ$n$\textgreater $\mathbf{F}_n$], along with a system message and a user query. This prompt structure enables the model to jointly reason about geometry, appearance, and contextual relationships.

The model is optimized end-to-end using the cross-entropy loss over response tokens. The training objective is formulated as
\begin{equation}
\mathcal{L}(\theta) = -\sum_{i=1}^{k} \log P\left(s_i^{\text{res}} \mid s_{[1,\dots,i-1]}^{\text{res}},\, s^{\text{prefix}}\right),
\end{equation}
where $k$ is the length of the response, $s_i^{\text{res}}$ is the generated target response, and $s_{[1,\dots,i-1]}^{\text{res}}$ is the previous $i-1$ tokens in the response. $\theta$ denotes the trainable parameters.

\section{Experiments}
\label{sec:experiments}

\begin{table*}[t]
\centering
\small
\setlength{\tabcolsep}{6pt} % reduce column spacing
\resizebox{\textwidth}{!}{
\begin{tabular}{lccccccccc}
\toprule
\textbf{Method} 
& \multicolumn{5}{c}{\textbf{ScanQA}} 
& \multicolumn{2}{c}{\textbf{SQA3D}} 
& \multicolumn{2}{c}{\textbf{ScanRefer}} \\
& BLEU-1 & BLEU-4 & METEOR & ROUGE & CIDEr 
& EM & EM-R 
& Acc@0.25 & Acc@0.5 \\
\midrule
\multicolumn{10}{l}{\textit{Expert Models}} \\
ScanRefer~\cite{chen2020scanrefer} & - & - & - & - & - & - & - & 37.3 & 24.3 \\
ScanQA~\cite{azuma2022scanqa} & 30.2 & 10.1 & 13.1 & 33.3 & 64.9 & - & - & - & - \\
SQA3D~\cite{ma2022sqa3d} & - & - & - & - & - & 46.6 & - & - & - \\
3DJCG~\cite{cai20223djcg} & - & - & - & - & - & - & - & 49.6 & 37.3 \\
3D-VLP~\cite{jin2023context} & 30.5 & 11.1 & 13.5 & 34.5 & 67.0 & - & - & 51.4 & 39.5 \\
M3DRef-CLIP~\cite{zhang2023multi3drefer} & - & - & - & - & - & - & - & 51.9 & 44.7 \\
3D-VisTA~\cite{zhu20233d} & - & 13.1 & 13.9 & 35.7 & 72.9 & 48.5 & - & 50.6 & 45.5 \\
ConcreteNet~\cite{unal2023three} & - & - & - & - & - & - & - & 50.6 & 46.5 \\
PQ3D~\cite{zhu2024unifying} & 43.0 & - & 17.8 & - & 87.8 & 47.1 & - & - & 51.2 \\
\midrule
\multicolumn{10}{l}{\textit{LLM-based Models}} \\
LAMM (Vicuna-13B)~\cite{yin2023lamm} & - & 5.8 & - & - & 42.4 & - & - & - & 3.4 \\
Chat-3D (Vicuna-7B)~\cite{wang2023chat} & 29.1 & 6.4 & 11.9 & 28.5 & 53.2 & - & - & - & - \\
ZSVG3D (GPT-4)~\cite{yuan2024visual} & - & - & - & - & - & - & - & 36.4 & 32.7 \\
3D-LLM (BLIP-2-FlanT5)~\cite{hong20233d} & 39.3 & 12.0 & 14.5 & 35.7 & 69.4 & - & - & 30.3 & - \\
LL3DA (OPT-1.3B)~\cite{chen2024ll3da} & - & 13.5 & 15.9 & 37.3 & 76.8 & - & - & - & - \\
Grounded 3D-LLM (Tiny-Vicuna-1B)~\cite{chen2024grounded} & - & 13.2 & - & - & 75.9 & - & - & 48.6 & 44.0 \\
LEO (Vicuna-7B)~\cite{huang2023embodied} & - & 11.5 & 16.2 & 39.3 & 80.0 & 50.0 & 52.4 & - & - \\
Scene-LLM (LLaMA-2-7B)~\cite{fu2024scene} & 43.6 & 12.0 & 16.6 & 40.0 & 80.0 & 54.2 & - & - & - \\
3DGraphLLM (Vicuna-7B)~\cite{zemskova20243dgraphllm} & - & 12.1 & - & - & 87.6 & 53.1 & - & 57.0 & 51.3 \\
Chat-Scene (Vicuna-7B)~\cite{huang2024chat} & 43.2 & 14.3 & 18.0 & 41.6 & 87.7 & 54.6 & 57.5 & 55.5 & 50.2 \\
\textbf{Descrip3D (Ours, Vicuna-7B)} & \textbf{44.0} & \textbf{14.5} & \textbf{18.6} & \textbf{43.1} & \textbf{93.7} & \textbf{55.7} & \textbf{58.4} & \textbf{57.2} & \textbf{51.8} \\
\bottomrule
\end{tabular}
}
\caption{Performance on ScanQA~\cite{azuma2022scanqa}, SQA3D~\cite{ma2022sqa3d}, and ScanRefer~\cite{chen2020scanrefer}. “Expert models” are tailored for specific tasks using task-oriented heads, while “LLM-based models” are designed for general instructions and responses. \textbf{Descrip3D achieves the highest performance across all benchmarks}, outperforming prior LLM-based methods by leveraging relational textual descriptions through a dual-level integration for more precise and context-aware reasoning.}
\label{tab:scanqa-sqa3d-scanrefer}
\end{table*}

\begin{figure*}
  \centering
  \begin{subfigure}{0.7\linewidth}
    \centering
    \includegraphics[width=0.95\linewidth]{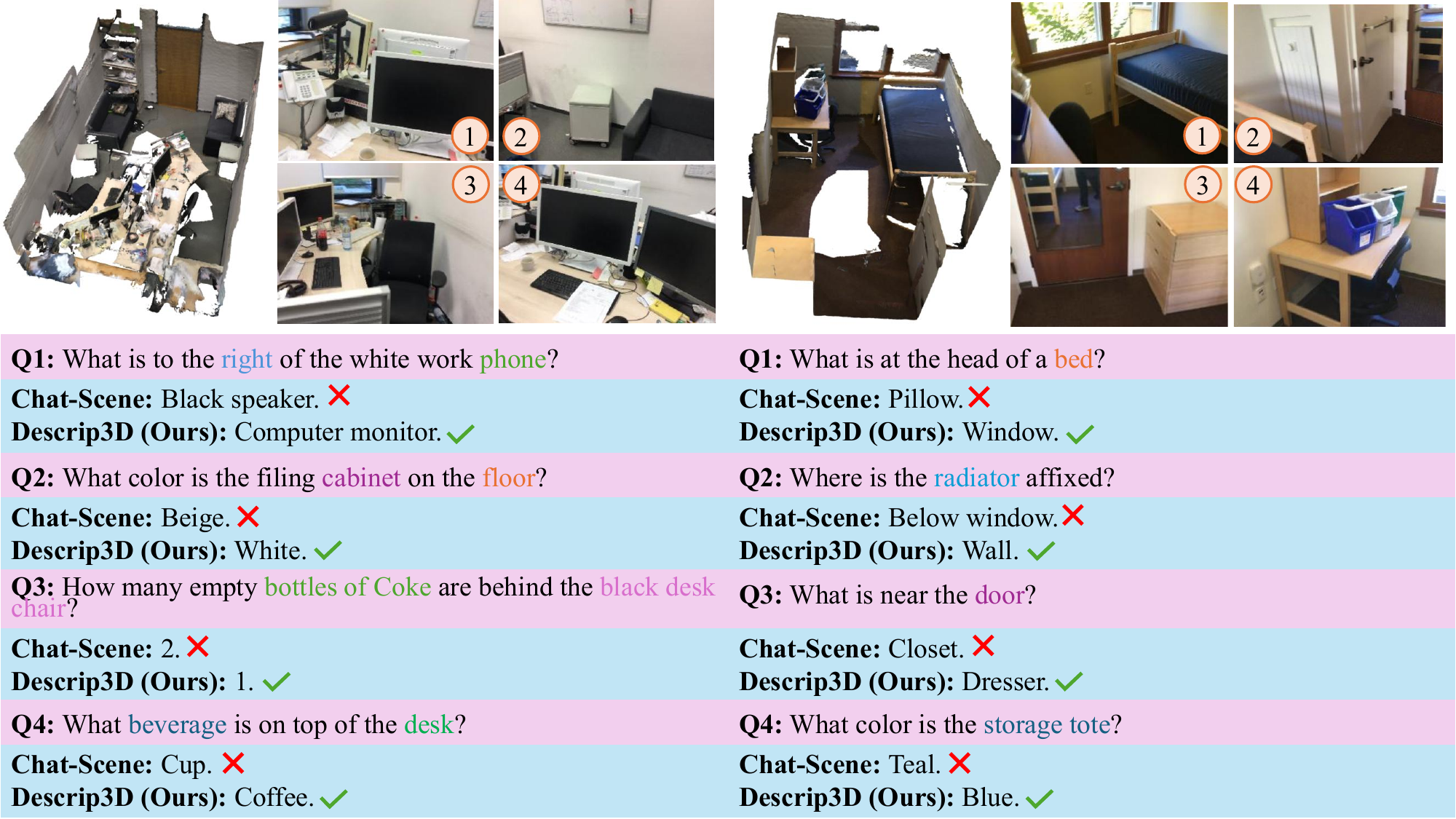}
    \caption{Qualitative Comparison of Question Answering.}
    \label{fig:qualitative_qa}
  \end{subfigure}
  \hfill
  \begin{subfigure}{0.28\linewidth}
    \centering
    \includegraphics[width=0.8\linewidth]{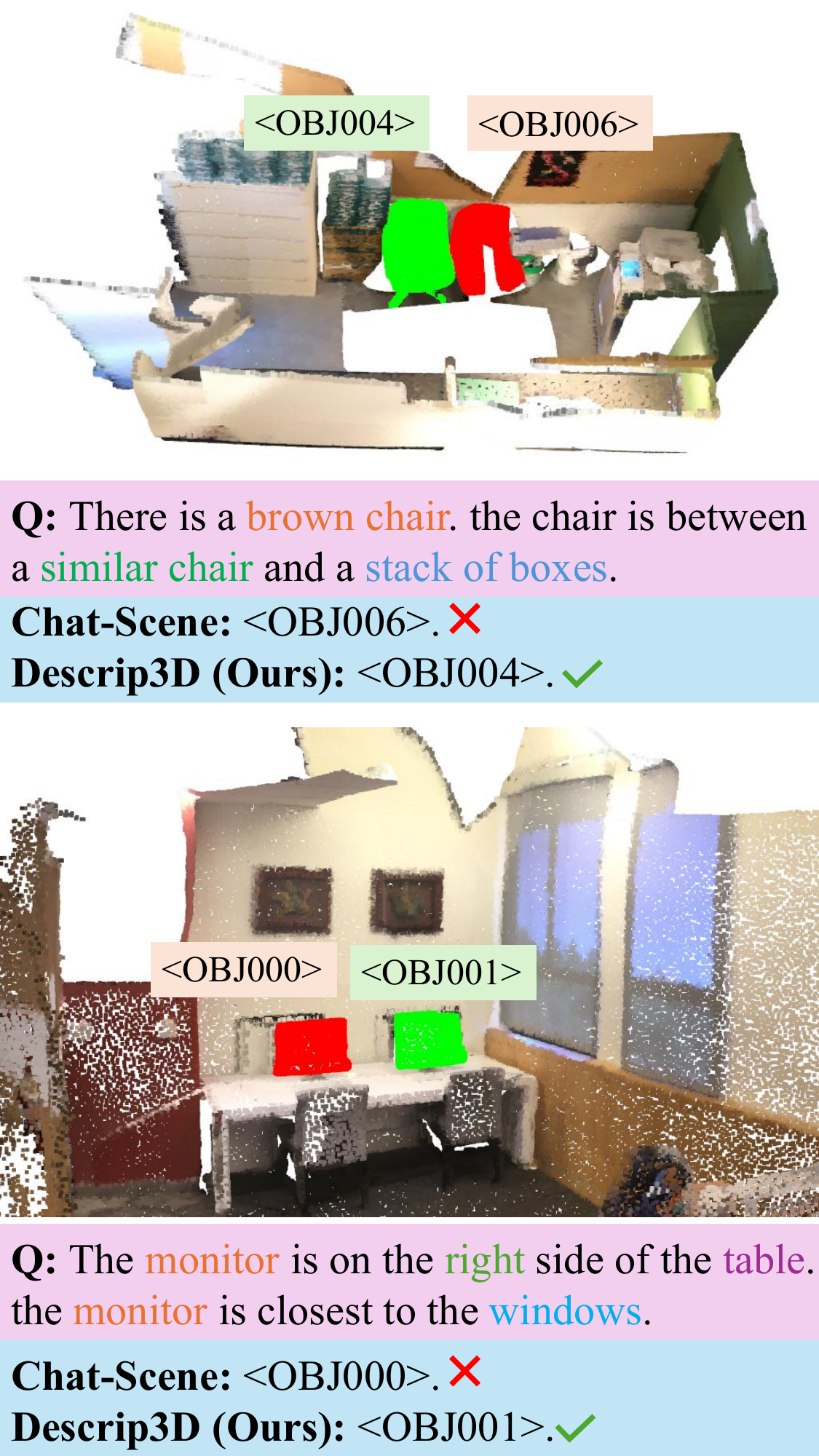}
    \caption{Qualitative Comparison of Grounding.}
    \label{fig:qualitative_grounding}
  \end{subfigure}
  \caption{Qualitative comparison of 3D scene understanding tasks. \textbf{Descrip3D outperforms Chat-Scene, especially in cases involving complex spatial grounding or multi-object reasoning}, due to its use of a dual-level integrated relational textual descriptions that enhance contextual understanding.}
  \label{fig:qualitative}
\end{figure*}

%-------------------------------------------------------------------------
\subsection{Datasets and Metrics}

\paragraph{Datasets.} 
To evaluate generalizability, we conduct experiments on five benchmarks spanning four major 3D vision-language tasks: ScanRefer~\cite{chen2020scanrefer} for single-object grounding, Multi3DRefer~\cite{zhang2023multi3drefer} for compositional multi-object grounding, Scan2Cap~\cite{chen2021scan2cap} for 3D-aware captioning, and ScanQA~\cite{azuma2022scanqa} plus SQA3D~\cite{ma2022sqa3d} for scene-level QA. All are derived from ScanNet~\cite{dai2017scannet}, a richly annotated dataset of 1,513 indoor scenes with point clouds, RGB images, and camera poses. This shared backbone provides consistent semantics across tasks and enables multi-task evaluation. In our experiments, we strictly follow the official splits provided by each benchmark (ScanRefer, Multi3DRefer, Scan2Cap, ScanQA, and SQA3D). All are built on ScanNet, where splits are defined at the scene level, ensuring disjoint train/val/test sets. Thus, evaluation scenes and objects are never partially seen during training, eliminating data leakage.

\paragraph{Metrics.} 
We use standard evaluation protocols from prior work~\cite{huang2024chat}. For ScanRefer~\cite{chen2020scanrefer}, we report thresholded accuracies Acc@0.25 and Acc@0.5, which assess whether the predicted object bounding box has an IoU with the ground truth exceeding 0.25 or 0.5. For Multi3DRefer~\cite{zhang2023multi3drefer}, which involves grounding multiple targets, we use the F1 score at IoU thresholds of 0.25 and 0.5. For the captioning task Scan2Cap~\cite{chen2021scan2cap}, we adopt CIDEr@0.5 and BLEU-4@0.5, integrating captioning quality with spatial alignment via IoU. For visual question answering, ScanQA~\cite{azuma2022scanqa} is evaluated using CIDEr and BLEU-4, while SQA3D~\cite{ma2022sqa3d} is evaluated using Exact Match (EM) and its refined variant EM-R as proposed in LEO~\cite{huang2023embodied}. 

\begin{table*}[ht]
\centering
\small
\setlength{\tabcolsep}{6pt} % reduce column spacing
\renewcommand{\arraystretch}{0.7}
\begin{tabular}{cccccccc}
\toprule
\multirow{2}{*}{\textbf{3D}} & 
\multirow{2}{*}{\textbf{2D}} & 
\multirow{2}{*}{\textbf{Text}} &
\textbf{ScanRefer} & 
\textbf{Multi3DRef} & 
\textbf{Scan2Cap} & 
\textbf{ScanQA} & 
\textbf{SQA3D} \\
& & & Acc@0.5 & F1@0.5 & C@0.5 & CIDEr & EM \\
\midrule
\cmark & \xmark & \xmark & 42.0 & 44.3 & 63.5 & 86.5 & 52.7 \\
\xmark & \cmark & \xmark & 44.0 & 48.1 & 68.4 & 86.2 & 52.4 \\
\xmark & \xmark & \cmark & 2.0 & 8.2 & 16.1 & 73.2 & 49.7 \\
\xmark & \cmark & \cmark & 46.7 & 49.5 & 73.6 & 89.4 & 53.8 \\
\cmark & \xmark & \cmark & 41.5 & 44.9 & 63.3 & 86.9 & 53.1 \\
% \cmark & \cmark & \xmark & 51.2 & 54.6 & 75.4 & 90.0 & 54.9 \\
\cmark & \cmark & \xmark & 49.5 & 52.8 & 73.0 & 90.4 & 54.1 \\
\cmark & \cmark & \cmark & \textbf{51.8} & \textbf{55.1} & \textbf{77.2} & \textbf{93.7} & \textbf{55.7}  \\
\bottomrule
\end{tabular}
\caption{Modality ablation results across five benchmarks. Text-only features underperform due to lack of geometric grounding, while 2D and 3D features alone offer moderate performance. The best results are achieved when all three modalities are integrated, highlighting their complementary strengths.}
\label{tab:ablation_modalities}
\end{table*}

\begin{table}[ht]
\centering
\small
\setlength{\tabcolsep}{6pt} % reduce column spacing
\renewcommand{\arraystretch}{0.7}
\begin{tabular}{lcc}
\toprule
\textbf{Method} & \multicolumn{2}{c}{\textbf{Multi3DRefer}}  \\
                & F1@0.25 & F1@0.5 \\
\midrule
\multicolumn{3}{l}{\textit{Expert Models}} \\
3DVG-Transformer~\cite{zhao20213dvg}    & - & 25.5 \\
3DJCG~\cite{cai20223djcg}    & - & 26.6 \\
D3Net~\cite{chen2022d3net}    & - & 32.2 \\
M3DRef-CLIP~\cite{zhang2023multi3drefer}    & 42.8 & 38.4 \\
PQ3D~\cite{zhu2024unifying}    & - & 50.1 \\
\midrule
\multicolumn{3}{l}{\textit{LLM-based Models}} \\
% 3DGraphLLM     & 60.1 & 55.4 \\
Grounded 3D-LLM~\cite{chen2024grounded}  & 44.7 & 40.8 \\
Chat-Scene~\cite{huang2024chat}     & 57.1 & 52.4 \\
\textbf{Descrip3D (Ours)}  & \textbf{59.4} & \textbf{55.1} \\
\bottomrule
\end{tabular}
\caption{Performance on Multi3DRefer~\cite{zhang2023multi3drefer}. \textbf{Descrip3D achieves the best performance}, with notable gains in multi-object reasoning due to its dual-level relational text modeling.}
\label{tab:multi3drefer}
\end{table}

\begin{table}[ht]
\centering
\small
\setlength{\tabcolsep}{6pt} % reduce column spacing
\renewcommand{\arraystretch}{0.7}
\begin{tabular}{lcc}
\toprule
\textbf{Method} & \multicolumn{2}{c}{\textbf{Scan2Cap}}  \\
                & C@0.5 & B-4@0.5 \\
\midrule
\multicolumn{3}{l}{\textit{Expert Models}} \\
Scan2Cap~\cite{chen2021scan2cap}       & 35.2 & 22.4 \\
3DJCG~\cite{cai20223djcg}          & 49.5 & 31.0 \\
3D-VLP~\cite{jin2023context}         & 54.9 & 32.3 \\
3D-VisTA~\cite{zhu20233d}       & 66.9 & 34.0 \\
Vote2Cap-DETR++~\cite{chen2024vote2cap} & 67.6 & \textbf{37.1} \\
\midrule
\multicolumn{3}{l}{\textit{LLM-based Models}} \\
LL3DA~\cite{chen2024ll3da}          & 65.2 & 36.8 \\
LEO~\cite{huang2023embodied}            & 68.4 & 36.9 \\
Grounded 3D-LLM~\cite{chen2024grounded}  & 70.2 & 35.0 \\
%3DGraphLLM     & 81.2 & 36.3 \\
Chat-Scene~\cite{huang2024chat}     & 77.1 & 36.3 \\
\textbf{Descrip3D (Ours)}  & \textbf{77.2} & 34.5 \\
\bottomrule
\end{tabular}
\caption{Performance on Scan2Cap~\cite{chen2021scan2cap}. \textbf{Descrip3D achieves the best C@0.5 performance} by leveraging richer object-level relational context.}
\label{tab:scan2cap}
\end{table}

\subsection{Implementation Details}
We extract 100 object proposals per scene using the Mask3D~\cite{schult2022mask3d} segmentation model. For 3D geometry, we use Uni3D~\cite{zhou2023uni3d} to extract point embeddings $\mathbf{Z}^{v_p}_i$. For 2D appearance, we adopt DINOv2~\cite{oquab2023dinov2} to extract object-level embeddings $\mathbf{Z}^{v_f}_i$, consistent with prior work (e.g., ChatScene~\cite{huang2024chat}, 3DGraphLLM~\cite{zemskova20243dgraphllm}), to ensure fair comparison. While stronger encoders could raise performance, our relational-text modality is independent of the 2D backbone. For text, we utilize the LLaVA-v1.5-7B~\cite{liu2023llava} model to generate descriptions for each object. These object descriptions are then encoded using the all-mpnet-base-v2 model, a SentenceTransformer~\cite{reimers-2019-sentence-bert} based on MPNet~\cite{song2020mpnet} that captures rich contextual semantics for downstream fusion, resulting in the embeddings denoted as $\mathbf{Z}^{v_t}_i$. Each modality is projected to the token space using a three-layer MLP. 

We use Vicuna-7B-v1.5~\cite{chiang2023vicuna} as the base LLM to ensure a fair comparison with prior work (ChatScene, 3DGraphLLM, 3D-Ground-LLM). Grounded-3D-LLM primarily reports results with Tiny-Vicuna-1B, but also notes little difference with Vicuna-7B. We fine-tune it with LoRA~\cite{hu2022lora} (rank 16) for 3 epochs, batch size 32, lr $5\times10^{-6}$, cosine schedule. Training is conducted on two 80G NVIDIA A100 GPUs and completes in about 24 hours. We observe that training for 2 epochs yields improved performance on the ScanQA dataset, and we adopt this setting in relevant evaluations.

\subsection{Performance Comparison}
To comprehensively evaluate the effectiveness of our proposed relational text modality, we conduct extensive experiments across four representative 3D vision-and-language tasks: 3D visual grounding,  question answering, multi-object 3D visual grounding, and scene captioning. These tasks are benchmarked using the five widely adopted datasets mentioned before. 

\begin{table*}[ht]
\centering
\small
\setlength{\tabcolsep}{6pt} % reduce column spacing
\renewcommand{\arraystretch}{0.7}
\begin{tabular}{lccccc}
\toprule
\multirow{2}{*}{\textbf{Reference Style}} & \textbf{ScanRefer} & \textbf{Multi3DRefer} & \textbf{Scan2Cap} & \textbf{ScanQA} & \textbf{SQA3D} \\
 & Acc@0.5 & F1@0.5 & C@0.5 & CIDEr & EM \\
\midrule
Object Name Only         & 51.6 & 54.5 & 75.6 & 91.4 & 55.1 \\
Object Name + ID         & 51.1 & 53.9 & 74.1 & 91.7 & 54.2 \\
Object ID Only (Ours)    & \textbf{51.8} & \textbf{55.1} & \textbf{77.2} & \textbf{93.7} & \textbf{55.7} \\
\bottomrule
\end{tabular}
\caption{Ablation study on object reference style in prompt-level text injection. \textbf{Using object IDs alone yields the best performance across all benchmarks}, as it avoids ambiguity and improves alignment between queries and object descriptions.}
\label{tab:ablation-reference-style}
\end{table*}

\begin{table*}[ht]
\centering
\small
\setlength{\tabcolsep}{6pt} % reduce column spacing
\renewcommand{\arraystretch}{0.7}
\begin{tabular}{ccccccc}
\toprule
\multirow{2}{*}{\textbf{Embedding}} & \multirow{2}{*}{\textbf{Prompt}} &
\textbf{ScanRefer} & 
\textbf{Multi3DRef} & 
\textbf{Scan2Cap} & 
\textbf{ScanQA} & 
\textbf{SQA3D} \\
\textbf{} & & Acc@0.5 & F1@0.5 & C@0.5 & CIDEr & EM \\
\midrule
\xmark & \xmark & 50.2 & 52.4 & 77.1 & 87.7 & 54.6 \\
\cmark & \xmark & 50.9 & 54.7 & 76.4 & 93.6 & 55.2 \\
\xmark & \cmark & 51.6 & \textbf{55.1} & \textbf{77.3} & 89.9 & 54.7 \\
\cmark & \cmark & \textbf{51.8} & \textbf{55.1} & 77.2 & \textbf{93.7} & \textbf{55.7} \\
\bottomrule
\end{tabular}
\caption{Ablation study on text description and injection strategy. \textbf{Combining embedding-level and prompt-level injection consistently leads to the best overall performance}, demonstrating the complementary benefits of dual-level relational text integration.}
\label{tab:ablation_fusion_extended}
\end{table*}

\paragraph{Scene-level Question Answering.}
We evaluate our model on ScanQA~\cite{azuma2022scanqa} and SQA3D~\cite{ma2022sqa3d}, both of which require comprehensive scene-level understanding and precise object-grounded reasoning. As shown in \cref{tab:scanqa-sqa3d-scanrefer}, our method achieves the highest scores across all major metrics compared with previous expert models and LLM-based methods, highlighting the strength of our dual-level integrated relational descriptions in enhancing context understanding. Notably, CIDEr, which emphasizes content relevance, shows a significant gain of 5.9 over PQ3D and 6.0 over Chat-Scene, confirming that our model generates more informative and relevant answers. On the SQA3D dataset, our method attains the highest EM and EM-R, again outperforming existing models. The strong results across both benchmarks demonstrate the effectiveness of enriching 3D object representations with object-level textual descriptions. As a general-purpose model, our model consistently outperforms task-specific systems, demonstrating strong versatility without the need for task-dependent architectures. Compared to knowledge-graph-based approaches such as 3DGraphLLM, our model achieves better alignment with language and context, suggesting that lightweight textual descriptions offer a more direct and interpretable semantic grounding. Overall, the results confirm that injecting fine-grained natural language descriptions of objects into the scene representation, through dual-level integration, significantly enhances the LLM's ability to handle 3D visual question answering tasks.

\paragraph{3D Visual Grounding.}

We evaluate 3D visual grounding on the ScanRefer~\cite{chen2020scanrefer} dataset. As shown in \cref{tab:scanqa-sqa3d-scanrefer}, our method achieves the best performance among all methods. Compared to strong expert models like ConcreteNet and 3D-VisTA, our model surpasses them by 5.3 and 6.3 at the stricter 0.5 IoU threshold, respectively. These expert models are trained with tailored 3D architectures and task-specific objectives, highlighting the strength of our unified and generalizable approach. Among LLM-based methods, our model also clearly outperforms Chat-Scene and 3DGraphLLM. The improvement over Chat-Scene, which uses only 2D and 3D embeddings, validates the great benefit of incorporating relational object-level textual descriptions through a dual-level integration approach.

\paragraph{Multi‑Object 3D Visual Grounding.}
We evaluate our method on Multi3DRefer~\cite{zhang2023multi3drefer}, a challenging benchmark requiring models to resolve complex multi-object visual grounding within 3D scenes. Unlike traditional single-object grounding, this task demands fine-grained relational reasoning among multiple entities. As shown in \cref{tab:multi3drefer}, our method achieves the best performance, outperforming both expert models and LLM-based baselines such as Chat-Scene and Grounded 3D-LLM.

\paragraph{3D Captioning.}
On Scan2Cap~\cite{chen2021scan2cap}, which evaluates the generation of 3D-aware captions for objects in complex indoor scenes, our model obtains the highest CIDEr score as shown in \cref{tab:scan2cap}. This indicates that its captions are more informative and semantically aligned with ground truth. While BLEU-4 is slightly lower, this metric favors exact n-gram overlap and can penalize the diverse relational phrasing produced by our model. Importantly, unlike expert captioning systems tailored to this task, Descrip3D is trained as a unified model across multiple benchmarks yet still delivers strong captioning results. This shows that relational textual descriptions not only improve grounding and QA but also enhance semantic richness and generalization.

\paragraph{Qualitative results.} \cref{fig:qualitative} presents qualitative comparisons between our method and Chat-Scene on two representative tasks: 3D question answering (\cref{fig:qualitative_qa}) and 3D visual grounding (\cref{fig:qualitative_grounding}). For question answering, our model shows stronger ability to understand relational context and resolve spatial references, whereas Chat-Scene struggles more with object relationships. For visual grounding, our model more accurately identifies target objects based on dual-level integrated complex natural language descriptions involving appearance and relative position, while Chat-Scene often fails to disambiguate visually or semantically similar candidates. These results highlight the benefit of injecting detailed object-level descriptions into the language model through a dual-level approach, enabling more precise and interpretable multi-modal reasoning.

In summary, our method achieves strong and consistent results across five benchmarks, demonstrating its effectiveness across tasks.

\subsection{Ablation Studies}
To evaluate the impact of our object-level text descriptions and dual-level integration strategies, we perform ablations on key components of the model.

\paragraph{Effect of Individual Modalities.}
We ablate 3D, 2D, and text features as shown in \cref{tab:ablation_modalities}. Single modalities are limited: 3D captures geometry but lacks appearance and relations, 2D provides surface cues without spatial grounding, and text alone conveys semantics but cannot localize objects. Pairwise combinations yield complementary gains, while integrating all three produces the strongest results. This confirms that relational text is most effective when fused with both 2D and 3D features.

\paragraph{Effect of Object Reference Style in Prompt-Level Description.}
We test different ways of expressing objects in prompt-level text injection as shown in \cref{tab:ablation-reference-style}. Using raw object names leads to ambiguity when multiple instances share the same category. Adding IDs to names introduces redundancy without improvement. In contrast, using IDs alone consistently gives the best results, ensuring unambiguous alignment between queries and object descriptions across all benchmarks.

\paragraph{Effect of Object Descriptions and Fusion Location.}
We also study where textual descriptions are integrated, as shown in \cref{tab:ablation_fusion_extended}. Injecting them only as embeddings enriches object semantics and improves QA tasks. Injecting them only into prompts benefits tasks like Scan2Cap and Multi3DRefer. Combining both yields the best overall results across all benchmarks, with especially strong improvements on ScanQA. These findings show that embedding- and prompt-level strategies are complementary, and together provide the most robust improvements in both spatial grounding and semantic reasoning.

\section{Conclusion}
\label{sec:conclusion}

We propose Descrip3D, a simple yet powerful framework for 3D scene understanding that explicitly models inter-object relationships using object-level textual descriptions. By integrating these relational cues through a dual-level strategy, embedding fusion and prompt injection, our method enables more effective multi-object reasoning.

Experiments on five benchmarks show that Descrip3D consistently outperforms expert and LLM-based models, with ablations confirming the importance of dual-level integration. Our results highlight the strength of language as a medium for structured scene representation. As future work, we plan to explore end-to-end training that jointly learns object description generation and extend our approach to dynamic or outdoor environments.

\section{Acknowledgments}
\label{sec:acknowledgments}
This project was sponsored by the US DoD LUCI (Laboratory University Collaboration Initiative) fellowship and the US Army Research Laboratory. The authors also acknowledge the Center for Advanced Research Computing (CARC) at the University of Southern California for providing computing resources.

\bibliographystyle{plain}
\renewcommand\refname{Reference}
\bibliography{main}

\appendix

\section{Examples of Generated Descriptions}

\begin{figure*}[ht]
  \centering
  \includegraphics[width=1\linewidth]{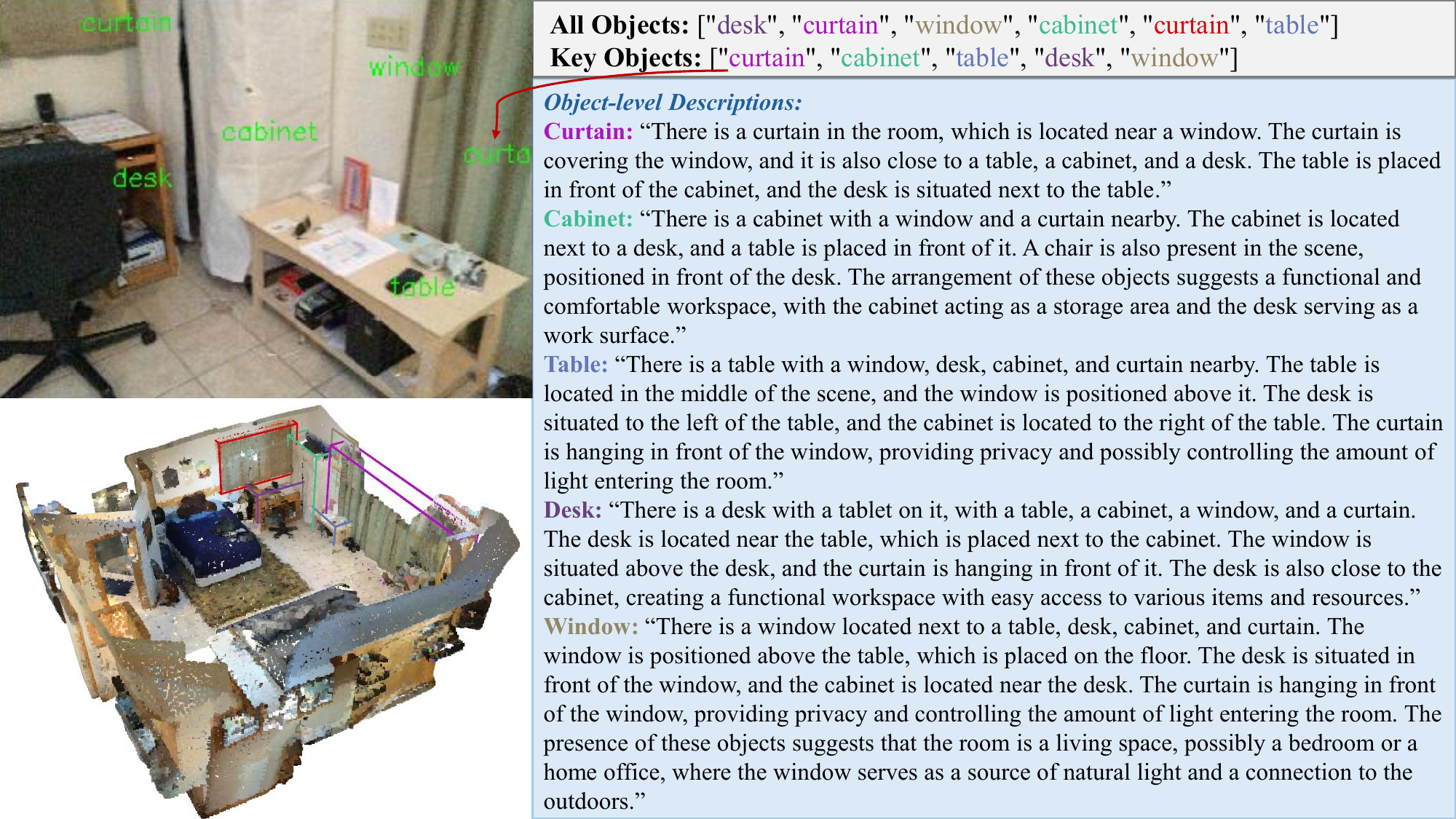}
  \caption{Qualitative examples of object-level relational descriptions generated using \textbf{Prompt A (Default)} with LLaVA-1.5}. The upper left part displays the image with object names, and the lower left shows the 3D bounding boxes. The right side contains relational descriptions emphasizing general proximity and object co-occurrence without enforcing strict spatial language.
  \label{fig:descript-example_prompt_A}
\end{figure*}

\begin{figure*}[ht]
  \centering
  \includegraphics[width=1\linewidth]{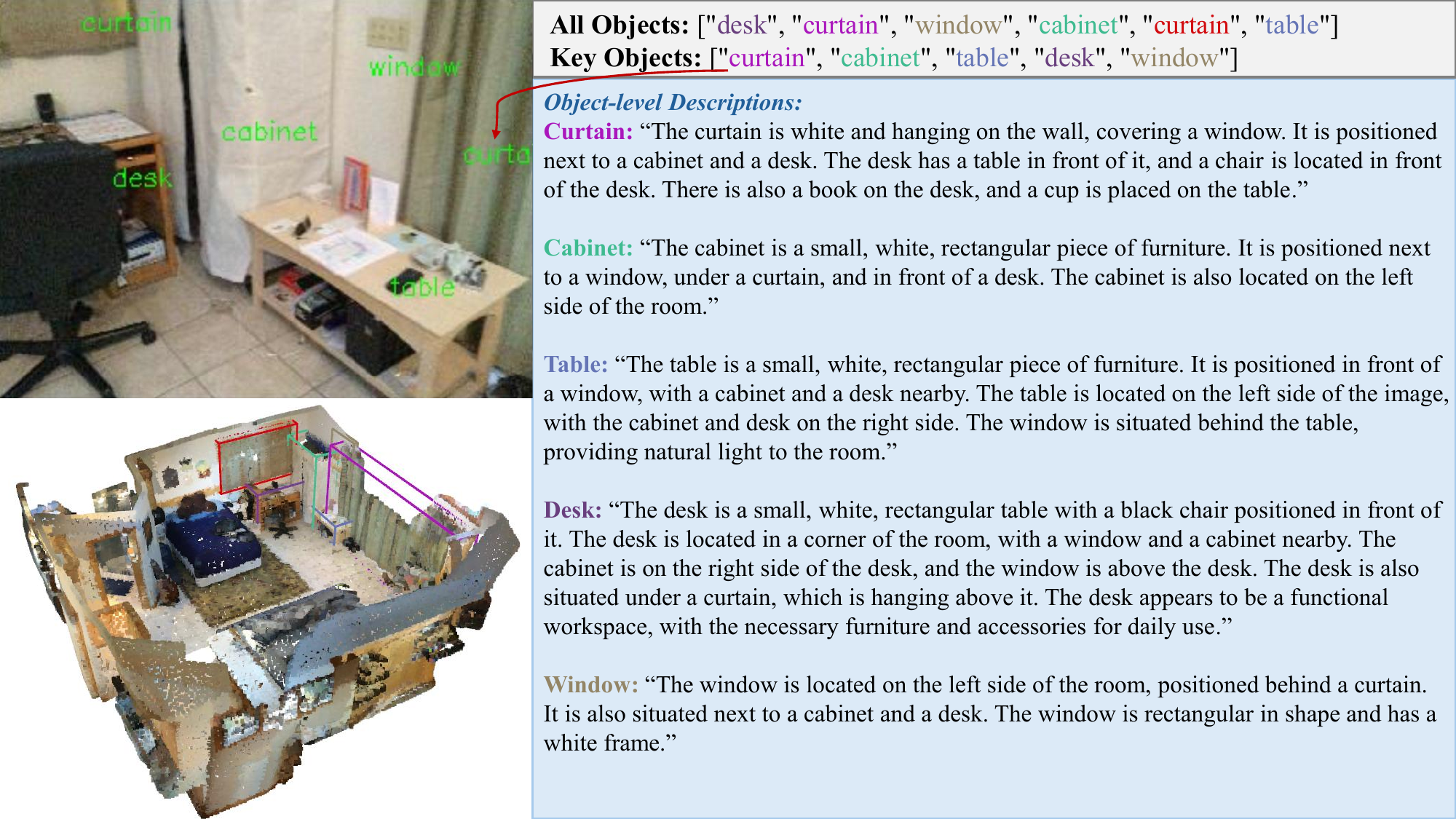}
  \caption{Qualitative examples of object-level relational descriptions generated using \textbf{Prompt B (Spatially Focused)} with LLaVA-1.5. Compared to Prompt A, these descriptions include more explicit spatial terms (e.g., “on the left,” “behind”) and visual attributes, resulting in shorter but more positionally grounded sentences.}
  \label{fig:descript-example_prompt_B}
\end{figure*}

We present qualitative examples of our object-level textual descriptions in ~\cref{fig:descript-example_prompt_A}. Starting from detected object proposals and their corresponding multi-view images, we overlay the object names at the center of the projection areas in the image, as illustrated in the upper left part of the example. We then generate relational descriptions using a vision-language model. Key objects, typically those centrally positioned in the scene, are selected as query anchors. For each key object, we prompt the model to describe its spatial relationships with all other detected objects, resulting in detailed, contextually grounded descriptions. The prompt used is: \textit{ ``Describe clearly and briefly the relationships between the \textless Key Object\textgreater in the scene and nearby objects (\textless Other Object 1\textgreater, \textless Other Object 2\textgreater, ..., \textless Other Object n\textgreater). Do not describe objects you cannot see.''} For example, the objects in the image are a desk, two curtains, a window, a cabinet, and a table. There are two curtains, but only the one on the right is considered a key object because the other is positioned at the edge of the image. The chosen curtain is described as covering the window and situated near the table, the cabinet, and the desk. These relational descriptions offer interpretable summaries of local neighborhoods and equip downstream models with structured scene understanding for improved reasoning.

\section{Ablation Study on Object Labels in Description Generation}
To examine the impact of explicitly overlaying object category names during relational description generation, we conduct an ablation study comparing two variants: one where multi-view images include projected object labels (ours), and one without. As shown in~\cref{tab:ablation-object-labels}, incorporating object labels consistently improves performance across all five benchmarks. The improvement is particularly notable in Scan2Cap and SQA3D, where more precise object references in the descriptions likely benefit caption generation and question answering. These results confirm that providing explicit category labels helps the vision-language model better ground each object and generate more informative relational descriptions.

\begin{table*}[ht]
\centering
\begin{tabular}{lccccc}
\toprule
\multirow{2}{*}{\textbf{Multi-view Image Input}} & \textbf{ScanRefer} & \textbf{Multi3DRefer} & \textbf{Scan2Cap} & \textbf{ScanQA} & \textbf{SQA3D} \\
 & Acc@0.5 & F1@0.5 & C@0.5 & CIDEr & EM \\
\midrule
Without Object Labels         & 51.5 & 54.8 & 75.6 & 93.5 & 54.6 \\
With Object Labels (Ours)    & \textbf{51.8} & \textbf{55.1} & \textbf{77.2} & \textbf{93.7} & \textbf{55.7} \\
\bottomrule
\end{tabular}
\caption{Ablation study on the effect of overlaying object category labels in multi-view images during relational description generation. \textbf{Adding object labels leads to consistent performance improvements across all benchmarks}, demonstrating their importance in guiding the vision-language model toward accurate grounding.}
\label{tab:ablation-object-labels}
\end{table*}

\section{Ablation Study on Prompt Design for Object-Level Descriptions}
To assess how different prompt formulations influence the quality of generated object-level relational descriptions and downstream 3D scene understanding, we compare two designs: a default prompt (Prompt A) that emphasizes relational conciseness, and a spatially grounded prompt (Prompt B) that encourages explicit spatial terms and appearance details.

\cref{fig:descript-example_prompt_A} presents qualitative examples generated using Prompt A (Default). This prompt directs the vision-language model (LLaVA-1.5) to describe relational context in a concise, human-like way, without placing heavy emphasis on precise spatial markers. As shown, descriptions tend to mention object co-occurrence and proximity in natural, readable sentences. For example, “The curtain is covering the window, and it is also close to a table, a cabinet, and a desk.” While this phrasing lacks precise positional anchoring, it aligns with how humans intuitively describe contextual relevance.

In contrast, \cref{fig:descript-example_prompt_B} illustrates the results of Prompt B (Spatially Focused). This prompt explicitly encourages the use of geometric relations (“on the left,” “in front of,” “behind”) and appearance details (“white,” “rectangular”), resulting in descriptions that are shorter but more spatially grounded. For instance, “The desk is located in a corner of the room… the window is above the desk… the cabinet is in front of the desk,” offers a clearer positional context but a less nuanced interpretation of function or co-usage.

To better understand how different prompts influence downstream model performance,~\cref{tab:prompt_comparison} provides the full text of each design, and~\cref{tab:ablation-prompt-design} summarizes quantitative results across five 3D vision-language tasks. While Prompt B encourages explicit spatial expressions (e.g., “to the left of,” “in front of”) and produces shorter sentences, it tends to focus narrowly on positional details, omitting functional or contextual cues. This results in descriptions that are more rigid but less informative overall. In contrast, Prompt A (Ours) generates a richer relational context with broader object co-occurrence and usage clues. As shown, Prompt A consistently outperforms Prompt B, suggesting that general, semantically rich descriptions better support multimodal reasoning than strictly spatial ones.

\begin{table*}[ht]
\centering
\small
\setlength{\tabcolsep}{6pt}
\renewcommand{\arraystretch}{1.2}
\begin{tabular}{p{0.2\linewidth} p{0.7\linewidth}}
\toprule
\textbf{Prompt Version} & \textbf{Prompt Text} \\
\midrule
\textbf{Prompt A (Default)} &
\texttt{Describe clearly and briefly the relationships between the <key\_object> in the scene and nearby objects (<other\_obj1>, <other\_obj2>, ...). Do not describe objects you cannot see. Do not describe green labels.} \\
\midrule
\textbf{Prompt B (Spatially Focused)} &
\texttt{Based on the image, describe both the appearance and spatial relationships of the <key\_object> in relation to nearby visible objects (<other\_obj1>, <other\_obj2>, ...). Include visual details like color, shape, size, or texture of the <key\_object>, and explain precisely how it is positioned relative to nearby visible objects (<other\_obj1>, <other\_obj2>, ...) using terms such as 'on the left', 'next to', 'under', 'in front of', 'behind', or 'on top of'. Only refer to what is clearly visible. Do not mention green text labels or objects not shown in the image."} \\
\bottomrule
\end{tabular}
\caption{Comparison of prompt designs used for generating object-level relational descriptions with LLaVA-1.5. Prompt A is our default, concise formulation emphasizing relational grounding. Prompt B explicitly encourages spatial terms (e.g., “left,” “in front of”) and detailed appearance cues.}
\label{tab:prompt_comparison}
\end{table*}

\begin{table*}[ht]
\centering
\begin{tabular}{lccccc}
\toprule
\multirow{2}{*}{\textbf{Prompt Design}} & \textbf{ScanRefer} & \textbf{Multi3DRefer} & \textbf{Scan2Cap} & \textbf{ScanQA} & \textbf{SQA3D} \\
 & Acc@0.5 & F1@0.5 & C@0.5 & CIDEr & EM \\
\midrule
Prompt B         & 51.4 & \textbf{55.1} & 74.1 & 92.3 & 55.2 \\
Prompt A (Ours)    & \textbf{51.8} & \textbf{55.1} & \textbf{77.2} & \textbf{93.7} & \textbf{55.7} \\
\bottomrule
\end{tabular}
\caption{Downstream performance using different prompts for generating object-level descriptions. Prompt B emphasizes spatial precision, while Prompt A (ours) encourages concise, general relational reasoning. Despite lacking explicit directional terms, Prompt A outperforms or matches Prompt B, suggesting that overly specific spatial descriptions may omit broader contextual signals useful for multimodal understanding.}
\label{tab:ablation-prompt-design}
\end{table*}

\section{Ablation Study on Description Generator Choice}

We further analyze the effect of the model used to generate object-level descriptions. While our main experiments adopt Vicuna-7B as the trainable backbone (to ensure fair comparison with prior methods such as Chat-Scene and 3DGraphLLM), the relational text can, in principle, be generated by any frozen captioner. \cref{tab:ablation-VLM} compares two options: generating descriptions with Vicuna-7B versus with LLaVA-1.5. Results show that replacing Vicuna with LLaVA-1.5 as the description generator improves downstream performance, especially on language-intensive tasks (e.g., +2.0 CIDEr on Scan2Cap, +2.1 CIDEr on ScanQA). This suggests that our framework is flexible with respect to the choice of description generator, and benefits from relation-dense captions produced by stronger multimodal models.

\begin{table*}[ht]
\centering
\begin{tabular}{lccccc}
\toprule
\multirow{2}{*}{\textbf{VLM}} & \textbf{ScanRefer} & \textbf{Multi3DRefer} & \textbf{Scan2Cap} & \textbf{ScanQA} & \textbf{SQA3D} \\
 & Acc@0.5 & F1@0.5 & C@0.5 & CIDEr & EM \\
\midrule
Vicuna-7B         & \textbf{51.8} & 55.0 & 75.2 & 91.6 & 55.6 \\
LLaVa-1.5 (Ours)    & \textbf{51.8} & \textbf{55.1} & \textbf{77.2} & \textbf{93.7} & \textbf{55.7} \\
\bottomrule
\end{tabular}
\caption{Impact of the choice of description generator. We compare object-level relational text produced by Vicuna-7B and by LLaVA-1.5 (used as frozen captioners). Using LLaVA-1.5 yields stronger downstream results.}
\label{tab:ablation-VLM}
\end{table*}

\section{Additional Qualitative Results and Failure Cases}

\begin{figure*}
  \centering
  \begin{subfigure}{0.7\linewidth}
    \centering
    \includegraphics[width=0.95\linewidth]{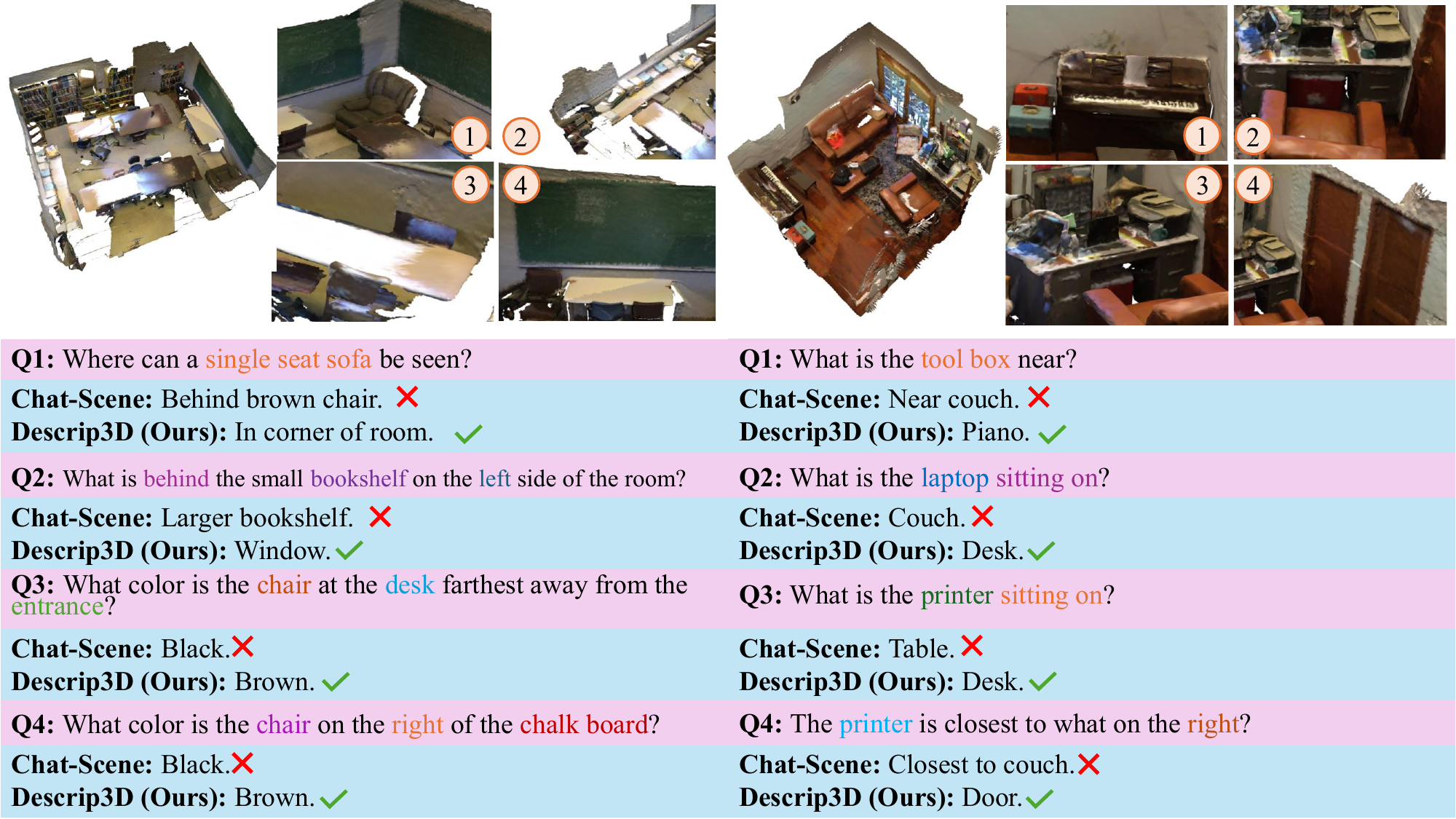}
    \caption{Additional qualitative comparison of question answering. Descrip3D correctly answers more challenging questions by leveraging precise spatial relations and context-aware relational cues from its textual descriptions.}
    \label{fig:add_qualitative_qa}
  \end{subfigure}
  \hfill
  \begin{subfigure}{0.28\linewidth}
    \centering
    \includegraphics[width=0.8\linewidth]{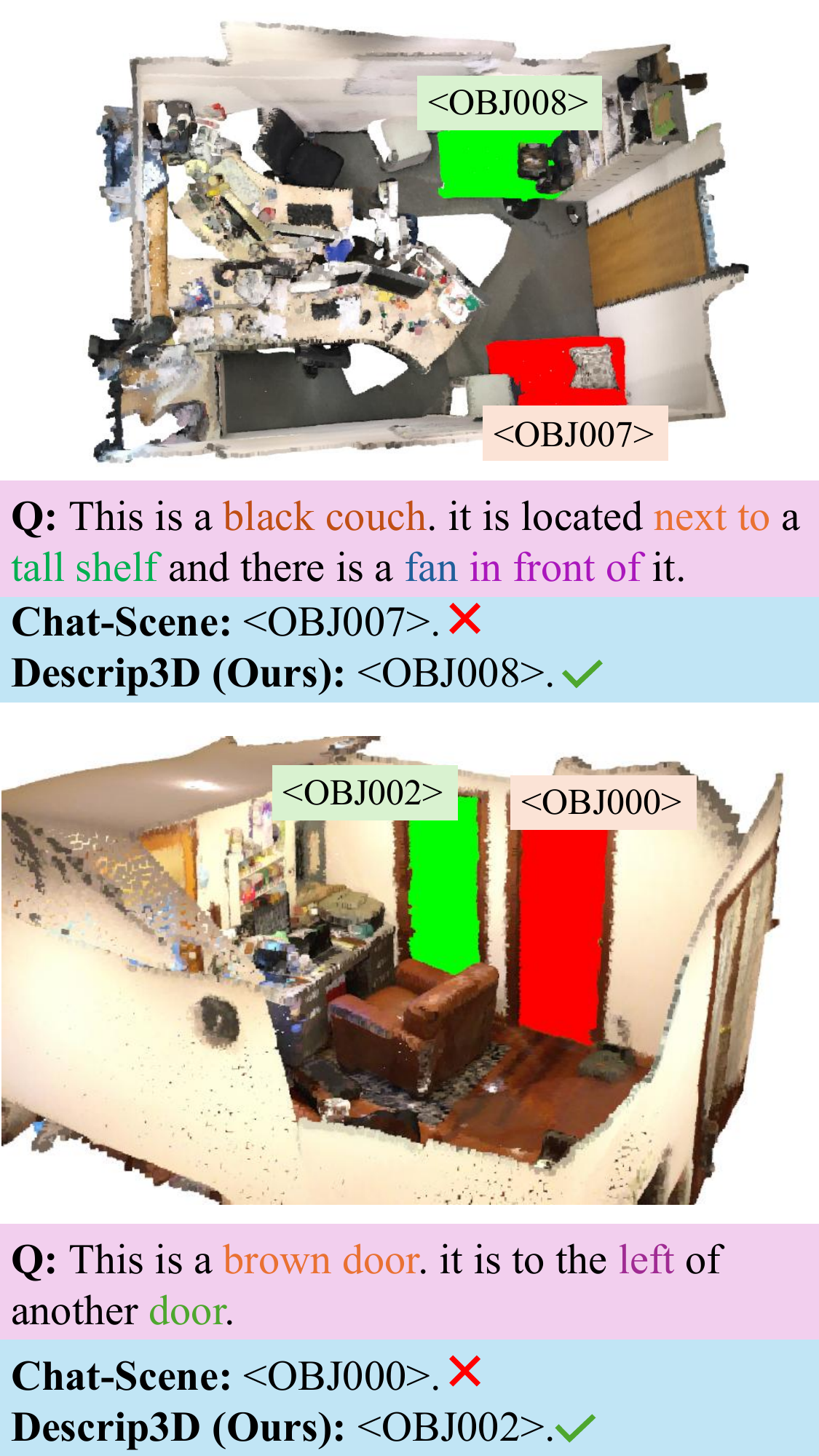}
    \caption{Additional qualitative comparison of grounding. Descrip3D provides more accurate object localization by resolving ambiguous object references using context-enhanced descriptions, such as "left of the fan" or “next to another door.”}
    \label{fig:add_qualitative_grounding}
  \end{subfigure}
  \caption{Additional Qualitative comparison of 3D scene understanding tasks. \textbf{Descrip3D outperforms Chat-Scene, especially in cases involving complex spatial grounding or multi-object reasoning}, due to its use of a dual-level integrated relational textual descriptions that enhance contextual understanding.}
  \label{fig:add_qualitative}
\end{figure*}

\paragraph{Additional Qualitative Examples}
To further demonstrate the strengths of Descrip3D, we present additional qualitative comparisons of both question answering and object grounding tasks in \cref{fig:add_qualitative}. In the QA task \cref{fig:add_qualitative_qa}, Descrip3D produces accurate answers in cases where Chat-Scene fails due to limited spatial awareness or insufficient contextual cues. For example, in the first question, while Chat-Scene incorrectly places the "single seat sofa" behind the brown chair, Descrip3D correctly identifies it as “in the corner of the room,” grounded by relational language. Similarly, Descrip3D succeeds in localizing queried objects such as the laptop and chairs based on complex object-to-object references, demonstrating its enhanced relational understanding. In the grounding task \cref{fig:add_qualitative_grounding}, Descrip3D resolves ambiguous references more reliably. For example, given a query like “the black couch next to a tall shelf and a fan,” Descrip3D identifies the correct object using spatial and contextual signals provided by the object descriptions. These results emphasize Descrip3D’s ability to perform robust reasoning in cluttered indoor environments where visual and geometric cues alone may be insufficient.

\begin{figure*}
  \centering
  \begin{subfigure}{0.7\linewidth}
    \centering
    \includegraphics[width=0.95\linewidth]{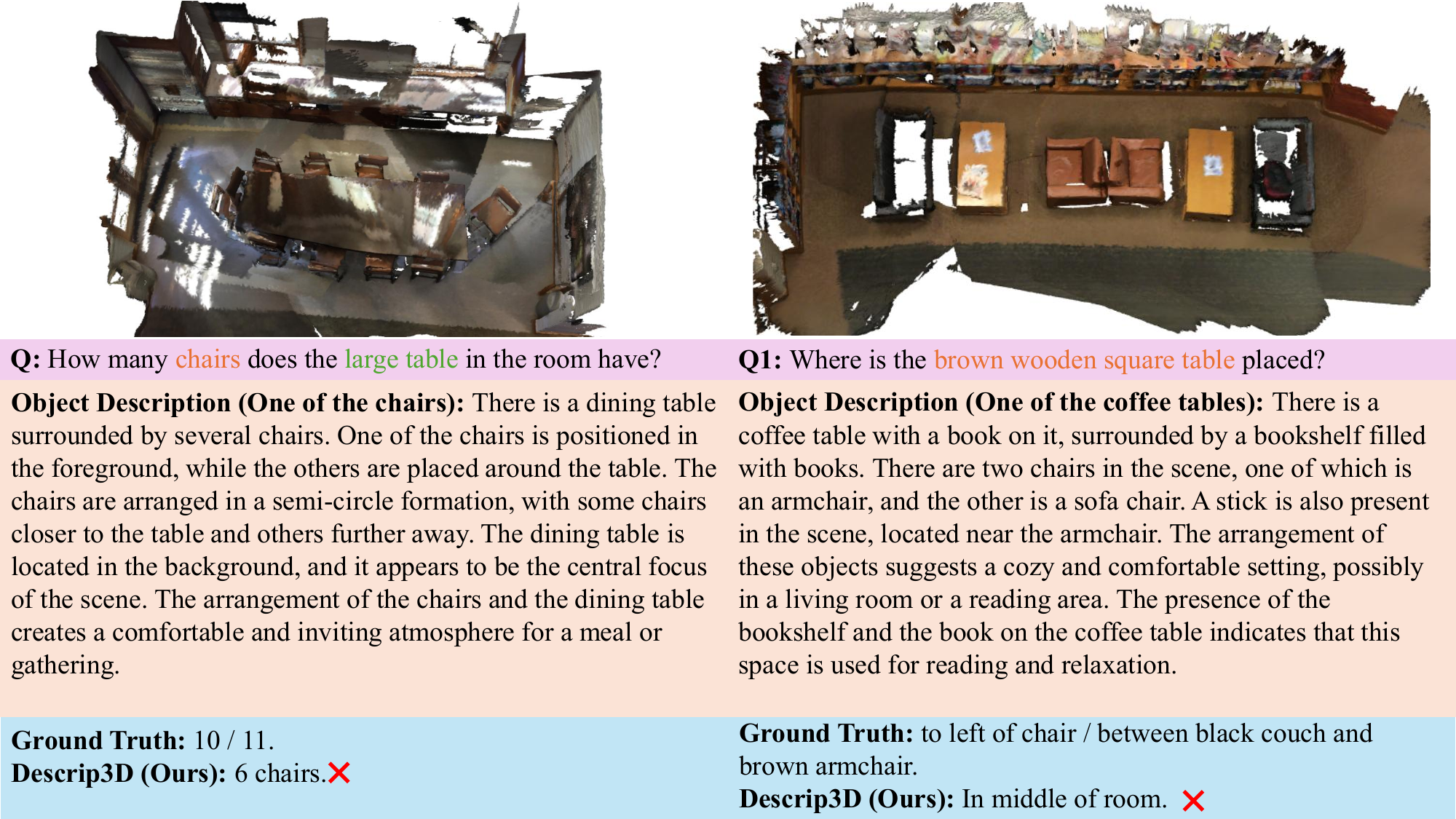}
    \caption{Additional qualitative comparison of question answering. Descrip3D fails in cases where the 3D detector misses objects or the query uses object names not aligned with detector output, limiting the effectiveness of textual reasoning.}
    \label{fig:fail_qualitative_qa}
  \end{subfigure}
  \hfill
  \begin{subfigure}{0.28\linewidth}
    \centering
    \includegraphics[width=0.8\linewidth]{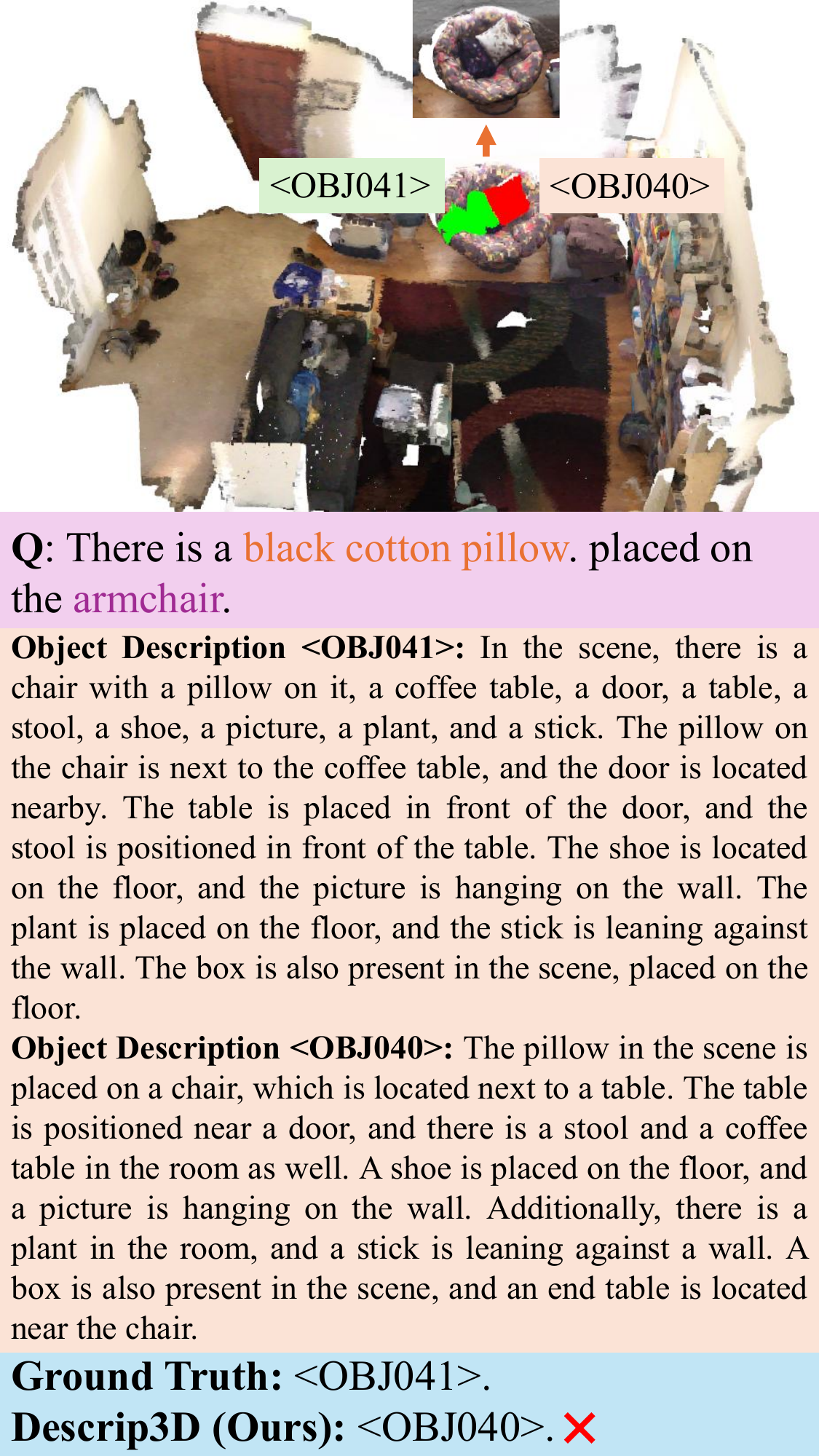}
    \caption{Additional qualitative comparison of grounding. Failures in grounding are often caused by incomplete object descriptions that omit key attributes (e.g., “black cotton pillow”), which are essential for accurate reference resolution.}
    \label{fig:fail_qualitative_grounding}
  \end{subfigure}
  \caption{Failure cases of 3D scene understanding tasks. While Descrip3D improves contextual reasoning, failure can still occur due to missing or ambiguous descriptions (e.g., color not mentioned) or mismatches between detection names and query expressions, especially in counting or spatial referencing.}
  \label{fig:fail_qualitative}
\end{figure*}

\paragraph{Failure Case Analysis}
Despite its improved performance, Descrip3D is not immune to errors. \cref{fig:fail_qualitative} illustrates several representative failure cases in both QA (\cref{fig:fail_qualitative_qa}) and grounding (\cref{fig:fail_qualitative_grounding}). In question answering, a common failure mode arises in counting tasks where performance is limited by upstream 3D detection accuracy. For instance, when the detector undercounts chairs around a table, Descrip3D cannot recover the correct answer solely through textual reasoning. Additionally, discrepancies between query phrasing (e.g., “square table”) and detector-generated object names (e.g., “coffee table”) introduce challenges in aligning language inputs with the available descriptions. In grounding, failures often occur when key attributes mentioned in the query (e.g., color or material) are missing or omitted in the generated object descriptions. As shown in ~\cref{fig:fail_qualitative_grounding}, the system fails to ground “the black cotton pillow” because the corresponding object description lacks explicit mention of its color, resulting in ambiguity during matching. These cases highlight limitations in both object detection accuracy and object description completeness, pointing to future directions for improving the robustness and coverage of relational grounding in 3D scene understanding.

\section{Additional Quantitative Results}
We evaluate our method using the standard metrics established in the original papers for each 3D scene-language dataset. To thoroughly assess the effectiveness of our approach, we perform extensive comparisons against a diverse set of baselines across multiple benchmarks. To complement the main results, we report additional evaluation metrics on the same datasets (ScanRefer, Multi3DRefer, and ScanQA) used in the main paper.  The results, summarized in \cref{tab:scanrefer-all} (ScanRefer), \cref{tab:multi3drefer-all} (Multi3DRefer), and \cref{tab:scanqa-all} (ScanQA), show our method consistently outperforms prior approaches across grounding and question answering tasks. On ScanRefer, Descrip3D achieves the highest overall accuracy. On Multi3DRefer, it leads in almost all grounding settings, with the best overall F1 scores. On ScanQA, it outperforms baselines in nearly all language metrics, including ROUGE-L, METEOR, and CIDEr. These results confirm the effectiveness of incorporating object-level textual descriptions through dual-level integration for 3D vision-language tasks.

\section{Prompt Template}
We adopt the same dialogue-style prompt format as Chat-Scene~\cite{huang2024chat}, consisting of a system message, a user instruction, and the corresponding assistant response. The system message sets the interaction context and introduces the object-level representation of the scene. Specifically, the scene is serialized as a flat sequence of object identifiers and features: [\textless OBJ001\textgreater $\mathbf{F}_1$ \textless OBJ002\textgreater $\mathbf{F}_2$ ... \textless OBJ$n$\textgreater $\mathbf{F}_n$], where $\mathbf{F}_i$ represents the feature embedding of the \(i\)th object. Each object identifier uniquely refers to a detected object in the scene. Users interact with the system by referencing these identifiers directly, and the assistant generates responses based on the identifiers. \cref{tab:prompt-template} provides an example of this prompt format.

\begin{figure*}[ht]
  \centering
  \includegraphics[width=0.9\linewidth]{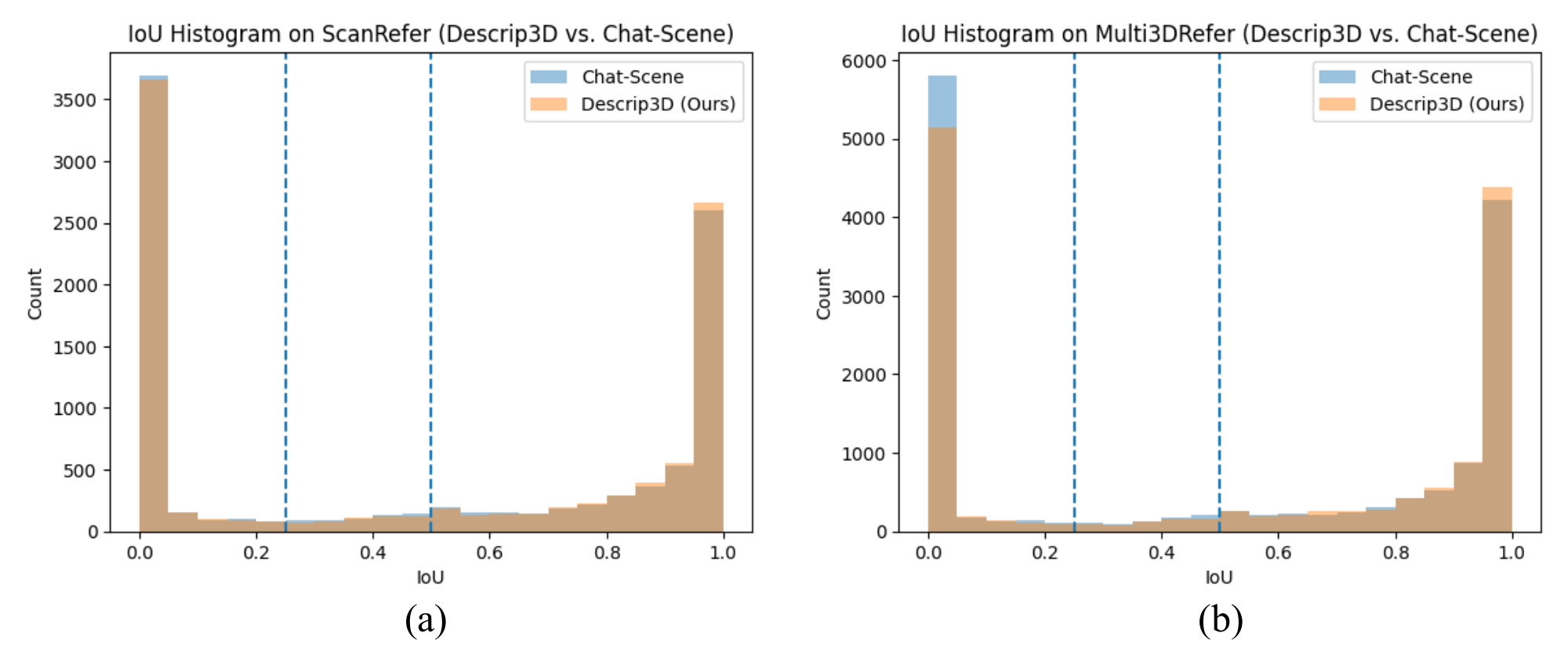}
  \caption{IoU distribution comparison between Chat-Scene and Descrip3D on (a) ScanRefer and (b) Multi3DRefer. The histograms show the per-sample IoU distributions, with dashed vertical lines indicating the 0.25 and 0.5 thresholds commonly used for visual grounding. Compared to Chat-Scene, Descrip3D consistently reduces the concentration of low-IoU cases and increases the density of high-IoU predictions, demonstrating more robust performance across the full distribution.}
  \label{fig:IoU_dist}
\end{figure*}

\begin{table}[ht]
\centering
\begin{tabular}{p{7cm}}
\toprule
\textbf{System:} A chat between a curious user and an artificial intelligence assistant. The assistant gives helpful, detailed, and polite answers to the user’s questions. The conversation centers around an indoor scene: [\textless OBJ001\textgreater $\mathbf{F}_1$ \textless OBJ002\textgreater $\mathbf{F}_2$...\textless OBJ$n$\textgreater $\mathbf{F}_n$]. \\
\textbf{User:} What is the ID of the object that matches the description "this is a brown chair. It is to the right of another chair near the end of the table."? [Generated description (may be noisy)] <OBJ002>: there is a <OBJ002> positioned next to a <OBJ003>... \\
\textbf{Assistant:} \textless OBJ002\textgreater \\
\bottomrule
\end{tabular}
\caption{Prompt template used during training and evaluation.}
\label{tab:prompt-template}
\end{table}

\begin{table*}[t]
\centering
\small
\setlength{\tabcolsep}{4pt}
\begin{tabular}{llcccccc}
\toprule
\textbf{Method} & \textbf{Venue} &
\multicolumn{2}{c}{\textbf{Unique}} &
\multicolumn{2}{c}{\textbf{Multiple}} &
\multicolumn{2}{c}{\textbf{Overall}} \\
& &
\textbf{Acc@0.25} & \textbf{Acc@0.5} &
\textbf{Acc@0.25} & \textbf{Acc@0.5} &
\textbf{Acc@0.25} & \textbf{Acc@0.5} \\
\midrule
ScanRefer~\cite{chen2020scanrefer}       & ECCV20   & 76.33 & 53.51 & 32.73 & 21.11 & 41.19 & 27.40 \\
TGNN~\cite{huang2021text}               & AAAI21   & 68.61 & 56.80 & 29.84 & 23.18 & 37.37 & 29.70 \\
SAT~\cite{yang2021sat}                  & ICCV21   & 73.21 & 50.83 & 37.64 & 25.16 & 44.54 & 30.14 \\
InstanceRefer~\cite{yuan2021instancerefer} & ICCV21   & 75.72 & 64.66 & 29.41 & 22.99 & 38.40 & 31.08 \\
3DVG-Transformer~\cite{zhao20213dvg}    & ICCV21   & 81.93 & 60.64 & 39.30 & 28.42 & 47.57 & 34.67 \\
MVT~\cite{huang2022multi}                  & CVPR22   & 77.67 & 66.45 & 31.92 & 25.26 & 40.80 & 33.26 \\
3D-SPS~\cite{luo20223d}                 & CVPR22   & 84.12 & 66.72 & 40.32 & 29.82 & 48.82 & 36.98 \\
ViL3DRel~\cite{chen2022language}        & NeurIPS22& 81.58 & 68.62 & 40.30 & 30.71 & 47.94 & 37.73 \\
3DJCG~\cite{cai20223djcg}                & CVPR22   & 83.47 & 64.34 & 41.39 & 30.82 & 49.56 & 37.33 \\
D3Net~\cite{chen2022d3net}              & ECCV22   & --    & 72.04 & --    & 30.05 & --    & 37.87 \\
BUTD-DETR~\cite{jain2022bottom}          & ECCV22   & 84.2  & 66.3  & 46.6  & 35.1  & 52.2  & 39.8 \\
HAM~\cite{chen2022ham}                  & ArXiv22  & 79.24 & 67.86 & 41.46 & 34.03 & 48.79 & 40.60 \\
3DRP-Net~\cite{wang20233drp}        & EMNLP23  & 83.13 & 67.74 & 42.14 & 31.95 & 50.10 & 38.90 \\
3D-VLP~\cite{jin2023context}               & CVPR23   & 84.23 & 64.61 & 43.51 & 33.41 & 51.41 & 39.46 \\
EDA~\cite{wu2023eda}                  & CVPR23   & 85.76 & 68.57 & \textbf{49.13} & 37.64 & 54.59 & 42.26 \\
M3DRef-CLIP~\cite{zhang2023multi3drefer}       & ICCV23   & 85.3  & 77.2  & 43.8  & 36.8  & 51.9  & 44.7 \\
3D-VisTA~\cite{zhu20233d}           & ICCV23   & 81.6  & 75.1  & 43.7  & 39.1  & 50.6  & 45.8 \\
ConcreteNet~\cite{unal2023three}  & ECCV24   & 86.40 & 82.05 & 42.41 & 38.39 & 50.61 & 46.53 \\
DORa~\cite{wu2024dora}                & ArXiv24  & --    & --    & --    & --    & 52.80 & 44.80 \\
Chat-Scene~\cite{huang2024chat}         & NeurIPS24  & 89.59 & 82.49 &  47.78 & 42.90 & 55.52 & 50.23 \\
\textbf{Descrip3D (Ours)}                           & --       & \textbf{90.79} & \textbf{83.23} &  \textbf{49.62} & \textbf{44.72} & \textbf{57.24} & \textbf{51.84} \\
\bottomrule
\end{tabular}
\caption{Performance comparison on the validation set of ScanRefer~\cite{chen2020scanrefer}.}
\label{tab:scanrefer-all}
\end{table*}

\begin{table*}[t]
\centering
\footnotesize
\setlength{\tabcolsep}{4pt}
\begin{tabular}{llcccccccccc}
\toprule
\textbf{Method} & \textbf{Venue} & 
\textbf{ZT w/o D} & \textbf{ZT w/ D} & 
\multicolumn{2}{c}{\textbf{ST w/o D}} & 
\multicolumn{2}{c}{\textbf{ST w/ D}} & 
\multicolumn{2}{c}{\textbf{MT}} & 
\multicolumn{2}{c}{\textbf{ALL}} \\
& & F1 & F1 & F1@0.25 & F1@0.5 & F1@0.25 & F1@0.5 & F1@0.25 & F1@0.5 & F1@0.25 & F1@0.5 \\
\midrule
3DVG-Trans+~\cite{zhao20213dvg} & ICCV21 & 87.1 & 45.8 & --   & --   & 16.7 & --   & 26.5 & --   & 25.5 & -- \\
D3Net (Grounding)~\cite{chen2022d3net} & ECCV22 & 81.6 & 32.5 & --   & --   & 23.3 & --   & 35.0 & --   & 32.2 & -- \\
3DJCG (Grounding)~\cite{cai20223djcg} & CVPR22 & \textbf{94.1} & 66.9 & --   & --   & 16.7 & --   & 26.2 & --   & 26.6 & -- \\
M3DRef-CLIP~\cite{zhang2023multi3drefer} & ICCV23 & 81.8 & 39.4 & 53.5 & 47.8 & 34.6 & 30.6 & 43.6 & 37.9 & 42.8 & 38.4 \\
Chat-Scene~\cite{huang2024chat} & NeurIPS24 & 90.3 & 62.6 & 82.9 & \textbf{75.9} & 49.1 & 44.5 & 45.7 & 41.1 & 57.1 & 52.4 \\
\textbf{Descrip3D (Ours)} & -- & 92.0 & \textbf{70.4} & \textbf{83.1} & \textbf{75.9} & \textbf{51.4} & \textbf{47.4} & \textbf{49.2} & \textbf{45.2} & \textbf{59.4} & \textbf{55.1} \\
\bottomrule
\end{tabular}
\caption{Performance comparison on the validation set of Multi3DRefer~\cite{zhang2023multi3drefer}.}
\label{tab:multi3drefer-all}
\end{table*}

\begin{table*}[t]
\centering
\small
\setlength{\tabcolsep}{6pt}
\begin{tabular}{llcccccccc}
\toprule
\textbf{Method} & \textbf{Venue} & \textbf{EM@1} & \textbf{B-1} & \textbf{B-2} & \textbf{B-3} & \textbf{B-4} & \textbf{ROUGE-L} & \textbf{METEOR} & \textbf{CIDEr}\\
\midrule
ScanQA~\cite{azuma2022scanqa}       & CVPR22   & 21.05 & 30.24 & 20.40 & 15.11 & 10.08 & 33.33 & 13.14 & 64.86\\
3D-VLP~\cite{jin2023context}           & CVPR22   & 21.65 & 30.53 & 21.33 & 16.67 & 11.15 & 34.51 & 13.53 & 66.97\\
3D-LLM~\cite{hong20233d}           & NeurIPS23 & 20.5  & 39.3  & 25.2  & 18.4  & 12.0  & 35.7  & 14.5  & 69.4\\
LL3DA~\cite{chen2024ll3da}          & CVPR24   & --    & --    & --    & --    & 13.53 & 37.31 & 15.88 & 76.79\\
LEO~\cite{huang2023embodied}                & ICML24   & --    & --    & --    & --    & 11.5  & 39.3  & 16.2  & 80.0\\
Scene-LLM~\cite{fu2024scene}    & WACV25  & \textbf{27.2} & 43.6 & 26.8  & 19.1  & 12.0  & 40.0  & 16.6  & 80.0\\
Chat-Scene~\cite{huang2024chat}     & NeurIPS24 & 21.62 & 43.20 & 29.06 & 20.57 & 14.31 & 41.56 & 18.00 & 87.70\\
\textbf{Descrip3D (Ours)}                       & --       & 22.67 & \textbf{44.36} & \textbf{30.51} & \textbf{22.08} & \textbf{15.70} & \textbf{43.01} & \textbf{19.06} & \textbf{93.71}\\
\bottomrule
\end{tabular}
\caption{Performance comparison on the validation set of ScanQA~\cite{azuma2022scanqa}.}
\label{tab:scanqa-all}
\end{table*}

\section{IoU Variance and Distribution}
To provide a more fine-grained view of grounding robustness beyond thresholded metrics (Acc@0.25, Acc@0.50), we report IoU distributions and variance. \cref{fig:IoU_dist} shows per-sample IoU histograms for ScanRefer and Multi3DRefer, both following the typical bimodal pattern with concentrations near IoU = 0 (failure) and IoU = 1 (successful grounding). Compared to Chat-Scene, Descrip3D consistently reduces the density of near-zero IoUs and increases the high-IoU mass. The measured variance is stable across datasets: ScanRefer variance $\approx$ 0.19 (std $\approx$ 0.44) and Multi3DRefer variance $\approx$ 0.20 (std $\approx$ 0.44). These results demonstrate that improvements are not driven by a handful of outlier cases but reflect consistent robustness across samples.
\end{document}